\documentclass[twoside]{article}
\usepackage[preprint]{aistats2027}

\usepackage[round]{natbib}

\usepackage[hyphens]{url}
\usepackage{graphicx}
\usepackage{algorithm}
\usepackage{booktabs}
\usepackage{tabularx}
\usepackage{algpseudocode}
\usepackage{xcolor}

\definecolor{revisionred}{RGB}{190,0,0}
\definecolor{feedbackgreen}{RGB}{0,100,0}
\definecolor{suggestionblue}{RGB}{0,51,153}
\newcommand{\annotationtag}[1]{%
  \if\relax\detokenize{#1}\relax\else
    \textsuperscript{\textbf{[AAAI~#1]}}\,%
  \fi
}

\newcommand{\parhead}[1]{\textbf{#1}\hspace{0.5em}}

\newcommand{\isarxivpreprint}{}

\usepackage{amsmath,amsfonts,bm}

\def\eqref#1{equation~\ref{#1}}

\def\1{\bm{1}}

\DeclareMathAlphabet{\mathsfit}{\encodingdefault}{\sfdefault}{m}{sl}
\SetMathAlphabet{\mathsfit}{bold}{\encodingdefault}{\sfdefault}{bx}{n}

\def\sR{{\mathbb{R}}}

\newcommand{\KL}{D_{\mathrm{KL}}}

\DeclareMathOperator*{\argmin}{arg\,min}

\usepackage{amsmath}
\usepackage{amssymb}
\usepackage{mathtools}

\newcommand{\method}{LADDER}

\providecommand{\bm}[1]{\boldsymbol{#1}} 

\newcommand{\vect}[1]{\bm{#1}}
\newcommand{\mat}[1]{\bm{#1}}

\newcommand{\W}{\mat{W}}
\newcommand{\X}{\mat{X}}

\newcommand{\norm}[1]{\left\lVert #1 \right\rVert}
\newcommand{\normF}[1]{\left\lVert #1 \right\rVert_{\mathrm{F}}}

\newcommand{\Lcal}{\mathcal{L}}      

\newcommand{\M}{\mat{M}}             
\newcommand{\odotop}{\odot}          

\newcommand{\supp}{\mathrm{supp}}

\newcommand{\Llayers}{L}             
\newcommand{\layer}{\ell}            
\newcommand{\WL}[1]{\W_{#1}}         
\newcommand{\XL}[1]{\X_{#1}}         
\newcommand{\ML}[1]{\M_{#1}}         
\newcommand{\WtL}[1]{\widetilde{\W}_{#1}} 
\newcommand{\El}[1]{\err_{#1}}       
\newcommand{\Sglob}{S}               
\newcommand{\Ccal}{\mathcal{C}}      
\newcommand{\Scand}{\mathcal{S}}     
\newcommand{\sprof}{\vect{s}}        

\newcommand{\err}{\mathcal{E}}       

\begin{document}

\runningauthor{Cuadrado \'Avila, Navarrete, Tak\'a\v{c}, Horv\'ath}
\twocolumn[
\aistatstitle{Building a Nested LADDER for LLM Unstructured Post-Training Pruning}
\aistatsauthor{Nicolas M. Cuadrado \'Avila \And Ivo Gollini Navarrete
  \AND Martin Tak\'a\v{c} \And Samuel Horv\'ath}
\aistatsaddress{Mohamed bin Zayed University of Artificial Intelligence, Abu Dhabi, United Arab Emirates}
]

\begin{abstract}
Non-uniform allocation can improve post-training pruning of large language models, yet quality can still degrade sharply at high sparsity.
To control degradation as sparsity increases, we introduce \method{} (\textbf{L}ayer-Wise \textbf{A}llocation via \textbf{D}amage-\textbf{D}riven \textbf{E}valuation and \textbf{R}efinement), which allocates sparsity incrementally.
At each step, it evaluates pruning an additional fraction of weights from each eligible layer and commits the layer update that causes the least end-to-end degradation per removed parameter.
This process produces non-uniform sparsity profiles and exactly nested masks across multiple budgets.
We evaluate \method{} on five OPT, Llama, and Qwen models across $10$--$90\%$ sparsity, using SparseGPT as its candidate-generation backend and comparing against uniform and adaptive allocation baselines.
Results reveal a clear operating-regime trade-off: \method{} is comparable or slightly weaker at benign sparsity but becomes substantially more robust from $70\%$ onward.
This high-sparsity advantage largely persists after posterior distillation across four of five models and translates into improved retention of aggregate zero-shot accuracy.
\end{abstract}

\begin{figure*}[t]
    \centering
    \includegraphics[width=0.989\textwidth]{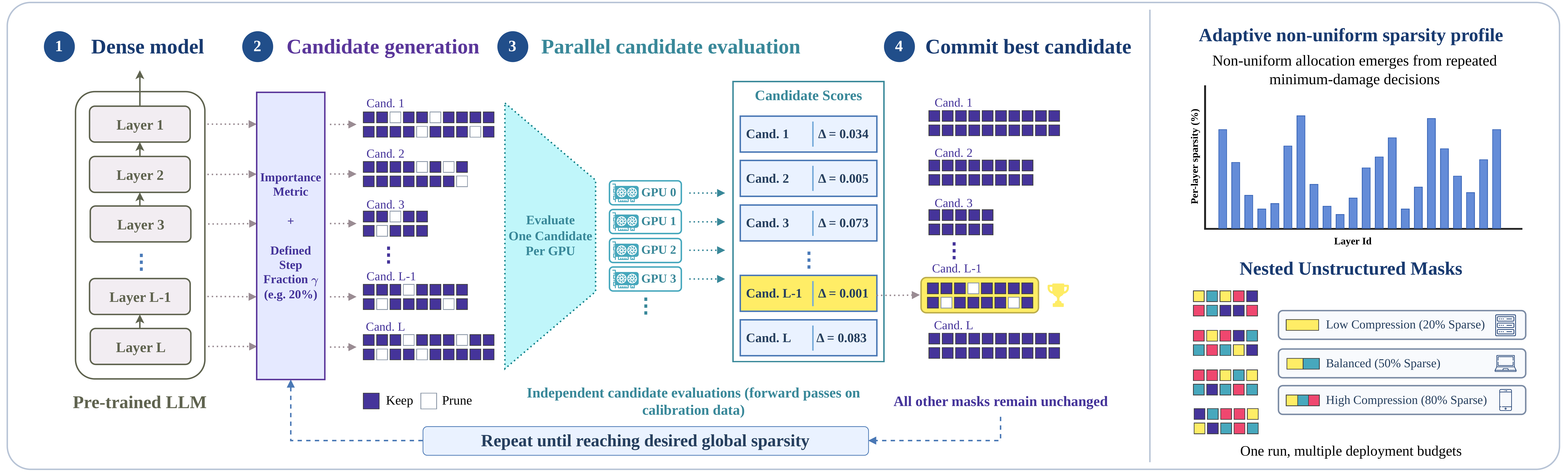} 
    \caption{
    Overview of \method{}.
    At each iteration, each eligible layer proposes a candidate pruning update using a fixed pruning step and a chosen local pruning backend.
    Each candidate is evaluated on calibration data, and the candidate with the least degradation per removed parameter is committed, producing an exactly nested mask trajectory.
    In our main experiments, the local backend is SparseGPT.
    }
\label{fig:visual_abstract}
\end{figure*}
\section{Introduction}

The strong performance of large language models (LLMs) has generally been attributed to parameter scaling, data volume, and increasing compute \citep{kaplanScalingLawsNeural2020}.
However, as the costs of training and deploying large-scale systems rise rapidly, some innovations become accessible only to well-funded organizations \citep{cottierRisingCostsTraining2025}.
For small-scale systems, inference faces increasingly stringent constraints on memory, bandwidth, and latency, particularly on edge devices \citep{quMobileEdgeIntelligence2025}.
At the same time, we observe a growing availability of open pre-trained models, including the Qwen series \citep{yangQwen3TechnicalReport2025}, Kimi \cite{MoonshotaiKimiK26Hugging2026}, and DeepSeek \cite{deepseek-aiDeepSeekV4HighlyEfficient2026}.
As a result, post-training compression becomes more attractive as training a model from scratch becomes more restricted.

The ``train-large-then-compress'' approach offers benefits beyond deployment convenience.
\citet{changProvableBenefitsOverparameterization2021} shows that in high-dimensional settings, training an overparameterized model and then pruning it can outperform training a smaller model from scratch, even when the relevant features are known.
This supports overparameterization and pre-trained LLMs as a basis for creating efficient subnetworks.

Numerous techniques have been developed for post-training compression, including quantization \citep{micikeviciusFP8FormatsDeep2022}, knowledge distillation \citep{xuSurveyKnowledgeDistillation2024}, low-rank approximation \citep{dentonExploitingLinearStructure2014a}, and pruning \citep{chengSurveyDeepNeural2024}.
Within the latter approach, unstructured pruning is a well-studied approach that removes individual parameters without modifying the model architecture.

Recent advancements in sparse representations and kernels have translated unstructured pruning into practical efficiency.
Notably, the MACKO-SpMV kernel demonstrates that runtime and storage benefits improve as sparsity increases, reporting both memory reduction and enhanced inference speed for a 50\% sparse LLM \citep{mackoMACKOSparseMatrixvector2026}.
Alongside broader developments in GPU kernels \citep{spectorThunderKittensSimpleFast2024}, the findings suggest the real-life feasibility of aggressive unstructured pruning.

Early influential approaches for LLMs, such as SparseGPT \citep{frantarSparseGPTMassiveLanguage2023a} and Wanda \citep{sunSimpleEffectivePruning2023a}, proposed effective methods for one-shot pruning via local decisions about which weights to remove within each layer, under a uniform layer-wise sparsity target. 
Nevertheless, recent research has shown that a non-uniform allocation of pruning across layers can lead to better performance, as redundancy varies with depth \citep{tangDarwinLMEvolutionaryStructured2026, gaoMaximumRedundancyPruning2025a, leOptiPruneEffectivePruning2025}. 
For instance, \citet{sunCurseDepthLarge2026} demonstrated that some deep layers can be completely removed with minimal negative impact, a phenomenon they refer to as the \textit{Curse of Depth}.
Despite these advances, existing non-uniform methods deteriorate sharply under aggressive sparsity, motivating the search for methods that can mitigate the detrimental impact in the high-sparsity regime and push the Pareto frontier.

We introduce \method{}, a post-training wrapper that controls non-uniform sparsity allocation by repeatedly selecting the layer whose next pruning step causes the least end-to-end degradation.
This process preserves quality as sparsity increases and, as a structural consequence, produces nested mask supports across sparsity budgets.
We provide an overview in Figure \ref{fig:visual_abstract}.

Our main contributions are:
(1) We introduce a conservative approach to non-uniform allocation that assigns sparsity incrementally according to the marginal end-to-end degradation measured from the current model state.
(2) We instantiate this approach as \method{}, a greedy deterministic backend-agnostic protocol whose monotone pruning decisions additionally produce exactly nested mask supports across sparsity budgets.
(3) Across five OPT, Llama, and Qwen models, we identify an operating-regime trade-off: \method{} becomes substantially more robust under aggressive pruning, with this advantage largely surviving posterior distillation and translating into improved aggregate accuracy retention across seven zero-shot downstream tasks.

\section{Background} \label{sec:background}

For each prunable layer $\layer \in \{1,\dots,\Llayers\}$ in a pre-trained model, let $\WL{\layer} \in \sR^{d_{\mathrm{out},\layer} \times d_{\mathrm{in},\layer}}$ denote the dense parameter matrix, $\XL{\layer} \in \sR^{d_{\mathrm{in},\layer} \times n}$ the matrix of calibration activations, $\ML{\layer} \in \{0,1\}^{d_{\mathrm{out},\layer} \times d_{\mathrm{in},\layer}}$ a binary pruning mask, and $\WtL{\layer}$ the post-pruning parameters after any optional local compensation or reconstruction step.
We define the layer-local degradation functional
\begin{align}
    \textstyle
    \El{\layer}(\ML{\layer}, \WtL{\layer}; \XL{\layer})
    =
    \normF{\WL{\layer}\XL{\layer} - (\ML{\layer} \odotop \WtL{\layer})\XL{\layer}}^2 ,
    \label{eq:lw_degradation}
\end{align}
as a standardized notation.
Although not every method optimizes \(\El{\layer}\) directly, it provides a common notation to compare pruning methods through the choice of mask selection ($\ML{\layer}$), weight reconstruction ($\WtL{\layer}$), and sparsity allocation.

We distinguish between \emph{global} pruning, which distributes a total sparsity budget across layers, and \emph{local} pruning, which focuses on pruning a specific layer after assigning it a sparsity level.

\parhead{Global Pruning.}
This approach formulates pruning as a model-level optimization problem.
Given a target sparsity, the goal is to identify the subset of parameters whose removal minimizes the expected degradation in model performance, regardless of the layers to which those parameters belong.
A commonly used tractable surrogate approximates the end-to-end objective with the sum of layer-local reconstruction errors:
\begin{align*}
    \textstyle
    \min_{\{\ML{\layer}, \WtL{\layer}\}_{\layer=1}^{\Llayers}}
    \quad &
    \sum_{\layer=1}^{\Llayers}
    \El{\layer}(\ML{\layer}, \WtL{\layer}; \XL{\layer}) \\
    \text{subject to}
    \quad &
    \sum_{\layer=1}^{\Llayers} \norm{\ML{\layer}}_0 \le (1-\Sglob) N,
\end{align*}
where $n_\layer = d_{\mathrm{out},\layer} \times d_{\mathrm{in},\layer}$ is the number of parameters in a given layer, $N = \sum_{\layer=1}^{\Llayers} n_\layer$ the total parameters in a model, and $\Sglob \in [0,1)$ the target global sparsity.
The global problem becomes one of minimizing the degradation by choosing the optimal vector $\{s_\layer\}_{\layer=1}^{\Llayers}$, where $s_\layer$ is the realized local sparsity.

\parhead{Local Pruning.}
On the other hand, this approach starts with a fixed layer-wise sparsity $s_\layer$ and asks how to achieve that local budget with minimal damage.
We can express this problem as a constrained approximation problem:
\begin{align*}
    \textstyle
    \min_{\ML{\layer}, \WtL{\layer}}
    \quad &
    \El{\layer}(\ML{\layer}, \WtL{\layer}; \XL{\layer}) \\
    \text{subject to}
    \quad &
    \norm{\ML{\layer}}_0 \le (1-s_\layer)n_\layer.
\end{align*}
Because jointly optimizing the pruning mask and the surviving weight values is intractable, existing approaches typically address only one of these variables.

Score-based methods focus on finding the optimal mask $\ML{\layer}^{\star}$ by solving
\begin{align*}
    \textstyle
    \argmin_{\ML{\layer}:\,\norm{\ML{\layer}}_0 \le (1-s_\layer)n_\layer}
    \;
    \min_{\WtL{\layer}}
    \El{\layer}(\ML{\layer}, \WtL{\layer}; \XL{\layer}),
\end{align*}
using a tractable saliency ranking, without caring about the surviving weight values.
A good example are the inflentuial methods OBD \citep{lecunOptimalBrainDamage1989} and OBS \citep{hassibiOptimalBrainSurgeon1993} which use scalable second-order information, an idea that has carried into modern settings \citep{navarreteWhatScalableSecondOrder2026}.
In contrast, reconstruction-based methods focus on finding the optimal surviving weight values $\WtL{\layer}^{\star}$ conditional on a selected or progressively constructed mask:
\begin{align*}
    \textstyle
    \argmin_{\WtL{\layer}}
    \quad &
    \El{\layer}(\ML{\layer}^{\star}, \WtL{\layer}; \XL{\layer}) \\
    \text{subject to}
    \quad &
    \supp(\WtL{\layer}) \subseteq \supp(\ML{\layer}^{\star}).
\end{align*}
Rather than focusing on mask quality, this approach aims to reconstruct the surviving parameters to better maintain the activation dynamics before pruning.

\subsection{Pioneering Pruning Methods}

Given a prescribed layer-wise sparsity profile, magnitude pruning removes weights with the smallest absolute values \citep{hagiwaraSimpleEffectiveMethod1994,hanLearningBothWeights2015a}, while Wanda incorporates input-activation norms into this ranking \citep{sunSimpleEffectivePruning2023a}.
Both are score-based methods that leave surviving values unchanged.
SparseGPT instead uses approximate second-order information to select removals while reconstructing surviving weights to preserve layer outputs \citep{frantarSparseGPTMassiveLanguage2023a}.
Together, these methods serve as standard reference points for evaluating post-training unstructured pruning under prescribed layer-wise sparsity targets.

\subsection{Layer-Wise Pruning Allocation Methods}

Various studies emphasize that compression should be strategically allocated rather than uniformly across layers.
OPQ determines layer-wise pruning ratios and quantization levels from pretrained parameter distributions \citep{huOPQCompressingDeep2021}. DPFPS progressively adapts pruning ratios based on layer sensitivity during CNN training \citep{ruanDPFPSDynamicProgressive2021}.
For LLMs, OWL applies non-uniform layer sparsities based on activation-outlier ratios \citep{yinOutlierWeighedLayerwise2024}, while DSA optimizes sparsity allocations based on layer statistics \citep{liDiscoveringSparsityAllocation2024}.
AlphaPruning leverages heavy-tailed spectral metrics \citep{luAlphaPruningUsingHeavyTailed2024a}, and Lua-LLM focuses on unstructured sparsity allocation \citep{luLuaLLMLearningUnstructuredSparsity2025}.
LSA employs minimal linear reconstruction error for sparsity assignment \citep{yangLSALayerwiseSparsity2025}, whereas BESA learns blockwise allocations with a differentiable reconstruction objective \citep{xuBESAPruningLarge2024}.
EvoPress utilizes evolutionary search to optimize dynamic compression profiles, accounting for interdependencies among decisions \citep{sieberlingEvoPressAccurateDynamic2025}.
SparseLLM coordinates pruning to address error propagation across transformer layers, albeit with homogeneous sparsity assumptions \citep{baiSparseLLMGlobalPruning2024a}.
Lastly, OATS combines sparse and low-rank approximations using a specified compression profile \citep{zhangOATSOutlierAwarePruning2024}.

\subsection{Elastic Model Families}

A complementary line of work studies elastic neural networks, which construct nested subnetworks with shared parameters, enabling a single model to support multiple deployment budgets.
Slimmable Networks introduced runtime-adjustable widths with independent normalization layers \citep{yuSlimmableNeuralNetworks2018a}, while Universally Slimmable Networks extended this idea to support arbitrary widths \citep{yuUniversallySlimmableNetworks2019}.
More recently, these principles have been applied to adaptive-width and depth language models \citep{houDynaBERTDynamicBERT2020a}, heterogeneous federated learning \citep{horvathFjORDFairAccurate2021}, and nested Transformers such as MatFormer \citep{devvritMatFormerNestedTransformer2024} and Flextron \citep{caiFlextronManyinOneFlexible2024}.
Moreover, FlexRank demonstrated that nested models can be extracted from pre-trained models using low-rank decompositions \citep{zacconeFlexRankNestedLowRank2026a}.

\subsection{Sparse-Model Refinement and Knowledge Transfer}

Knowledge distillation has become a standard approach for recovering performance after pruning.
CAP transfers knowledge from dense models and intermediate pruning states, showing that progressively sparser models can provide effective supervision during iterative pruning \citep{xuDenseSparseContrastive2022}.
AST similarly combines post-pruning retraining with dense-teacher supervision to improve the quality of semi-structured sparse LLMs under limited data and computation budgets \citep{huangPruningLargeLanguage2025}.
All of these methods apply distillation for a single sparse endpoint.
\begin{algorithm}[t]
    \caption{\method{} single-target allocation}
    \label{alg:ladder}
    \begin{algorithmic}[1]
        \Require committed masks $\{\ML{\layer}\}$ and weights $\{\WL{\layer}\}$
        \Require calibration data
        \Require target $\Sglob$, local backend, $\gamma$, $z_{\min}$
        \State Set round counter $r \gets 0$
        \While{global sparsity $<\Sglob$}
            \State Evaluate the committed model to obtain $m_{\mathrm{base}}^{(r)}$
            \ForAll{prunable layers $\layer$}
                \State Compute $z_\layer^{(r)}$
                \If{$z_\layer^{(r)}>0$}
                    \State Build candidate $(\widetilde{\ML{\layer}}^{(r)},\WtL{\layer}^{(r)})$
                    \State Evaluate the candidate and get $q_\layer^{(r)}$
                \EndIf
            \EndFor
            \State Choose $\layer^\star\gets\argmin_{\layer:z_\layer^{(r)}>0}q_\layer^{(r)}$
            \State Commit $(\ML{\layer^\star},\WL{\layer^\star})\gets(\widetilde{\ML{\layer^\star}}^{(r)},\WtL{\layer^\star}^{(r)})$
            \State Increment $r \gets r+1$
        \EndWhile
        \State \Return $\{\ML{\layer},\WL{\layer}\}_{\layer=1}^{\Llayers}$
    \end{algorithmic}
\end{algorithm}

\begin{figure*}[tb]
    \centering
    \includegraphics[width=0.989\textwidth]{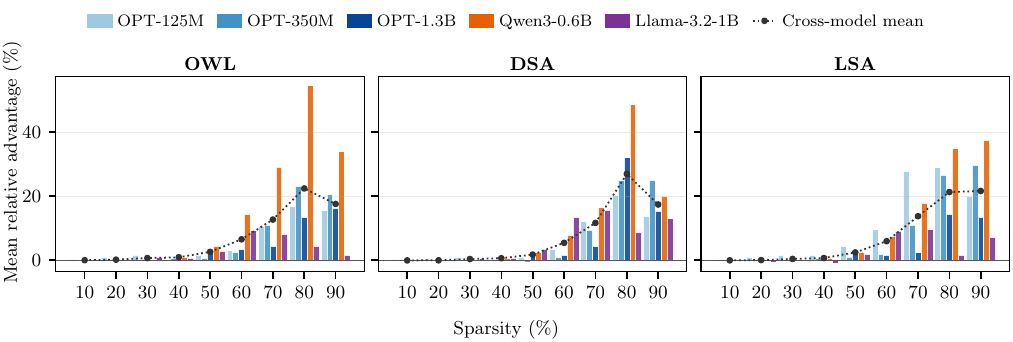}
    \caption{
        Pre-distillation relative loss advantage of \method{} over each baseline.
        The bars are the arithmetic mean of the relative advantages on WikiText-2, PTB, and C4.
        Positive values favor \method{}.}
    \label{fig:cross-model-sparsity-regimes}
\end{figure*}

\section{Methodology}

\subsection{Problem Setup}

We consider the layer-wise sparsity profile as the variable to be optimized. Let 
\begin{align*}
    \textstyle
    \sprof = (s_1, \dots, s_{\Llayers}) \in [0,1]^{\Llayers}
\end{align*}
represent the vector of local sparsities, and $\Sglob$ the desired global sparsity. We can formulate the allocation problem as follows:
\begin{align}
    \textstyle
    \min_{\sprof \in [0,1]^{\Llayers}}
    \Phi(\sprof)
    \quad
    \text{s.t.}
    \quad
    \sum_{\layer=1}^{\Llayers} n_\layer s_\layer = \Sglob N,
    \label{eq:ladder_base_formulation}
\end{align}
where $\Phi(\sprof)$ is the end-to-end calibration degradation induced by a layer-wise sparsity profile.
Here, $N$ counts only the parameters in the prunable linear layers, so $\Sglob$ is global sparsity over that declared pruning scope rather than over every model parameter.
Under a differentiable relaxation, any interior optimum, where $s_\layer^\star \in (0,1)$ for every layer, must satisfy the following first-order condition:
\begin{align*}
\textstyle
\frac{1}{n_\layer}\frac{\partial \Phi(\sprof^\star)}{\partial s_\layer} = \mu
\qquad
\text{for all prunable layers } \layer,
\end{align*}
for some shared multiplier $\mu$.
Thus, the marginal end-to-end damage per additional removed parameter must be equal across eligible layers.

Uniform sparsity satisfies this condition only when those per-parameter marginal damages happen to agree at the same local sparsity, which is generally not the case.
In practice, the sparsity profile is realized through binary keep-or-remove decisions, and pruning effects can interact across layers rather than vary smoothly or independently.
Consequently, the appropriate non-uniform profile generally cannot be determined from fixed, independently estimated layer costs.

\subsection{\underline{L}ayer-Wise \underline{A}llocation}

We introduce \method{}, which approximates the state-dependent allocation through controlled discrete updates, re-evaluating the available layer candidates after each committed step.
As shown in Figure \ref{fig:visual_abstract}, the process begins with the dense model and a per-layer score mask profile initialized to no pruning, represented as $\{ \ML{\layer}^{(0)} = \mathbf{1} \}_{\layer=1}^{\Llayers}$.

At round $r$, every prunable layer proposes a candidate mask $\widetilde{\ML{\layer}}^{(r)}$ that removes the $z_\layer^{(r)}$ least important parameters that survived from the current committed mask $\ML{\layer}^{(r)}$.
Layers for which $z_\layer^{(r)}=0$ will no longer generate candidates, allowing the candidate pool to diminish in subsequent rounds.
The candidate set for round $r$ is defined as follows:
\begin{align*}
    \textstyle
    \Ccal^{(r)} =
    \left\{\widetilde{\ML{\layer}}^{(r)}
    \mid 1\leq\layer\leq\Llayers,\ z_\layer^{(r)}>0\right\}.
\end{align*}
We evaluate all candidates using the same committed model state and a degradation criterion.
The committed mask profile is updated only with the one that shows the least degradation during each round. 
This repeats until reaching the global target sparsity $\Sglob$. 

\subsection{\underline{D}amage-\underline{D}riven \underline{E}valuation}

\parhead{Importance Metric.}
Our method uses existing local pruning methods as a backend to generate candidate masks.
Magnitude pruning and Wanda provide parameter scores, while SparseGPT additionally reconstructs the retained weights.

For state-dependent backends, the required statistics are refreshed from the current committed model: Wanda recomputes its activation-dependent scores, and SparseGPT updates its curvature information.
The candidate for layer $\layer$ at round $r$ is the pair $(\widetilde{\ML{\layer}}^{(r)},\WtL{\layer}^{(r)})$.
For score-only backends, $\WtL{\layer}^{(r)}=\WL{\layer}^{(r)}\odot\widetilde{\ML{\layer}}^{(r)}$.

The candidate's support must always be a subset of the committed support.
Reconstruction-based backends may alter surviving values, while masked coordinates stay zero.
Only the selected candidate remains.

Defining the number of surviving parameters in layer $\layer$ at round $r$ as \( K_\layer^{(r)} = \lVert\ML{\layer}^{(r)}\rVert_0 \), the remaining pruning budget is
\begin{align*}
    \textstyle
    R^{(r)}
    =
    \left\lfloor \Sglob N \right\rfloor
    -
    \sum_{\layer=1}^{\Llayers}
    \left(n_\layer-K_\layer^{(r)}\right),
\end{align*}
allowing us to define the candidate step for that layer:
\begin{align*}
    \textstyle
    z_{\layer}^{(r)}
    =
    \min\!\left(
        R^{(r)},
        \;K_{\layer}^{(r)},
        \;\max\!\left(z_{\min},\left\lceil \gamma n_\layer\right\rceil\right)
    \right).
\end{align*}
Here, $\gamma$ sets the nominal step as a fraction of the layer size, while $z_{\min}$ prevents excessively small updates.

\parhead{Degradation Criterion.}
We define $\Delta_\layer^{(r)}$ as the loss in model quality induced by a candidate:
\begin{align*}
\textstyle
    \Delta_\layer^{(r)} =
    \begin{cases}
        m_{\mathrm{cand}}^{(r,\layer)} - m_{\mathrm{base}}^{(r)}, & \text{for lower-is-better metrics}, \\
        m_{\mathrm{base}}^{(r)} - m_{\mathrm{cand}}^{(r,\layer)}, & \text{for higher-is-better metrics}.
    \end{cases}
\end{align*}
Here, $m_{\mathrm{base}}^{(r)}$ is the primary metric of the committed model and $m_{\mathrm{cand}}^{(r,\layer)}$ its value after temporarily activating the candidate for layer $\layer$.

Candidates are ranked by relative degradation per added zero,
\begin{align*}
    \textstyle
    q_\layer^{(r)} = \frac{\Delta_\layer^{(r)}}{z_\layer^{(r)}}.
\end{align*}
Ties are broken by raw degradation, then by the larger candidate step, and finally by stable layer order. For each round, the selected layer solves
\begin{align*}
\textstyle
\layer^\star = \argmin_{\layer:\,z_\layer^{(r)}>0} q_\layer^{(r)}.
\end{align*}
Algorithm \ref{alg:ladder} provides a detailed overview of our method.
The step fraction $\gamma$ dictates a straightforward trade-off: smaller steps lead to more precise allocation decisions but require more rounds to conclude.
Since candidates are assessed based on the current committed state, each decision incorporates the interactions evaluated up to that point.
However, the greedy trajectory may overlook options that could become more favorable after certain removals; thus, while exact nesting provides structural assurance, it does not guarantee optimality.

\begin{figure*}[tb]
    \centering
    \includegraphics[width=0.989\textwidth]{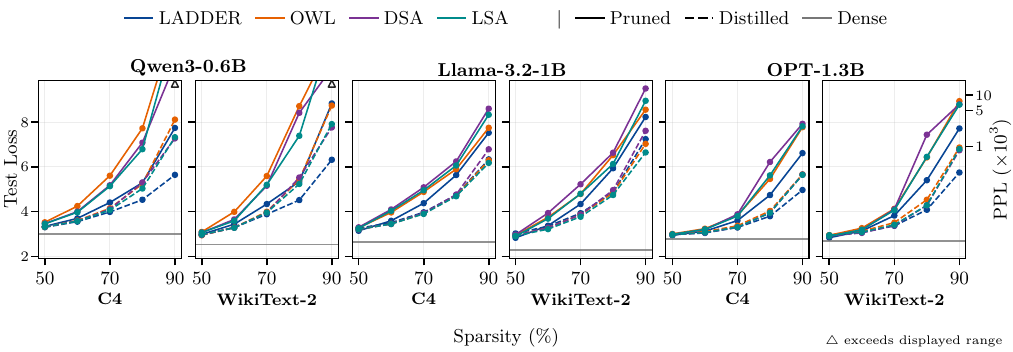}
    \caption{
        Language-modeling test loss before (solid) and after (dashed) posterior distillation on C4 and WikiText-2.
        Dataset panels are paired by model; the shared right axis gives the corresponding perplexity.}
    \label{fig:cross-model-loss-trajectories}
\end{figure*}

\subsection{\underline{R}efinement}

\parhead{Nestedness.}
We process requested sparsities in increasing order, initializing each run from the preceding checkpoint and restricting candidates to its support.
Consequently, for $s_a<s_b$, the resulting masks satisfy $M_{\layer}^{(b)}\subseteq M_{\layer}^{(a)}$.
Reconstruction-based backends may still change surviving weight values between checkpoints.
We empirically verify zero support violations.
The complete analysis is provided in Appendix~\ref{sec:appendix-diagnostics}.

\parhead{Posterior Distillation.}
Following prior work \citep{xuDenseSparseContrastive2022,huangPruningLargeLanguage2025}, we initialize one shared student from the pretrained dense model.
For each batch, all fixed sparsity masks are evaluated against the same dense-teacher outputs, their temperature-scaled KL losses are averaged, and one optimizer step updates the shared parameters.
We apply the same procedure to every baseline.

\begin{figure}[!ht]
    \centering
    \includegraphics[width=0.95\linewidth]{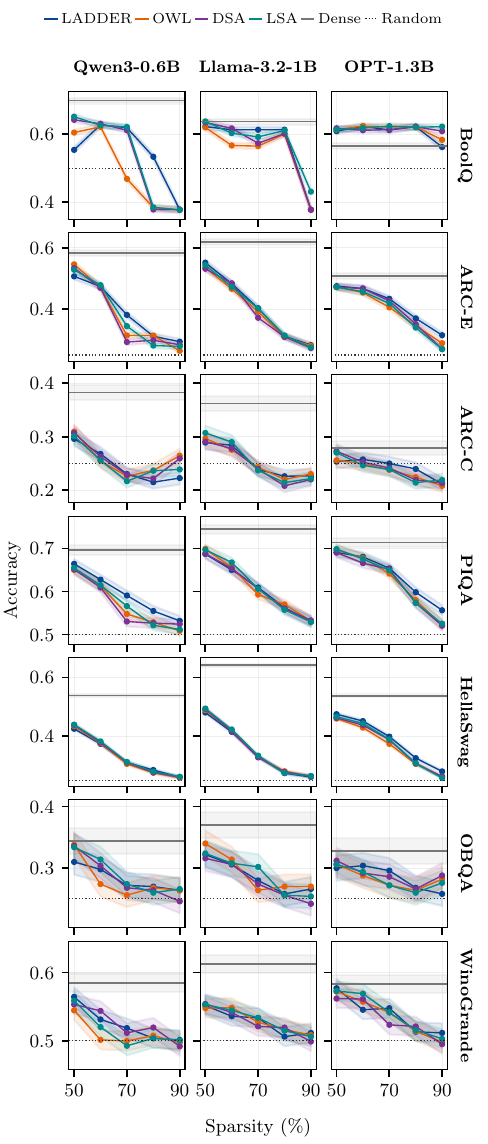}
    \vspace{-1.5em}
    \caption{
        Distilled zero-shot accuracy across seven downstream tasks.
        Rows share task accuracy limits across models.
        Shaded regions show the task standard errors reported by the evaluation harness.
        Horizontal references mark dense and random performance.
    }
    \label{fig:cross-model-distilled-downstream}
    \vspace{-1.5em}
\end{figure}

\section{Experiments \& Results}

Our experiments test whether continuously reassessing the least harmful pruning step is more beneficial as sparsification becomes more aggressive.
We first identify when this approach outperforms state-of-the-art non-uniform methods.
Then, we investigate if this advantage carries through the distillation process to downstream applications.
Lastly, we verify that \method{} produces nested masks and use ablations to study how our allocation interacts with different local pruning backends, and whether continuing from masks produced by other non-uniform methods can control their subsequent loss growth.

\subsection{Experimental Setup}

\parhead{Protocol.}
We evaluate OPT-125M, OPT-350M, OPT-1.3B \citep{zhangOPTOpenPretrained2022}, Qwen3-0.6B, and Llama-3.2-1B.
We use the standard language-modeling benchmarks WikiText-2, PTB, and C4 \cite{frantarSparseGPTMassiveLanguage2023a,sunSimpleEffectivePruning2023a}, to allow direct comparison with prior work.

Furthermore, we evaluate seven downstream tasks to test diverse zero-shot capabilities.
Following the pruning protocol from \citet{frantarSparseGPTMassiveLanguage2023a}, we prune all linear layers in the transformer blocks, excluding biases.
We evaluate requested global sparsities from 10\% to 90\% and focus the main comparisons on 50--90\%, where sparse kernels have shown practical gains \citep{mackoMACKOSparseMatrixvector2026}.

\parhead{Methods.}
We compare \method{} against the non-uniform allocation methods OWL, DSA, and LSA.
All methods use SparseGPT as a common local pruning backend, keeping local mask selection and weight reconstruction fixed to isolate the effect of the global allocation strategy.
All methods use the same 128 sequences of 2,048 tokens sampled from C4 for calibration; for \method{}, perplexity on this calibration sample additionally serves as the candidate-selection criterion.
Method-specific configurations and hyperparameters are provided in Appendix~\ref{sec:appendix-reproducibility}.

\parhead{Evaluation.}
We report corpus-level mean next-token cross-entropy and perplexity on WikiText-2, PTB, and C4, together with zero-shot accuracy on seven downstream tasks.
We additionally evaluate posterior distillation of the resulting sparsity-specific masks using a shared student initialized from the pretrained dense weights.
Complete language-modeling and distillation settings are provided in Appendix~\ref{sec:appendix-reproducibility}, while downstream evaluation details are given in Appendix~\ref{sec:appendix-downstream}.

\subsection{Language-Modeling Across Regimes}
\label{sec:experiments-operating-regime}

Figure \ref{fig:cross-model-sparsity-regimes} shows an operating-regime effect: the non-uniform baselines are often competitive at lower sparsity, results are mixed around 50--60\%, and \method{} separates most consistently from 70\% onward before distillation.
The bars preserve the magnitude of the model-level advantages across all five models, while Table \ref{tab:cross-model-pruned-lm-performance} reports the corresponding 70--90\% loss and perplexity values for the three models highlighted in the main figures. The complete five-model table is deferred to the Appendix \ref{sec:appendix-loss}.
\begin{table*}[tb]
    \center
    \caption{
        Pre-distillation language-modeling quality.
        Each entry reports test loss followed by perplexity in parentheses for each dataset.
        Lower is better, and bold indicates the best method for each model, dataset, and sparsity level.
    }
    \label{tab:cross-model-pruned-lm-performance}
    \resizebox{\textwidth}{!}{%
\begin{tabular}{llccccccccc}
\toprule
Model & Method & \multicolumn{3}{c}{70\%} & \multicolumn{3}{c}{80\%} & \multicolumn{3}{c}{90\%} \\
\cmidrule(lr){3-5} \cmidrule(lr){6-8} \cmidrule(lr){9-11}
 & & WT2 & PTB & C4 & WT2 & PTB & C4 & WT2 & PTB & C4 \\
\midrule
Qwen3-0.6B & LADDER & \textbf{4.34 (76.4)} & \textbf{5.00 (147.7)} & \textbf{4.40 (81.8)} & \textbf{5.34 (208.7)} & \textbf{6.26 (524.5)} & \textbf{5.29 (197.6)} & \textbf{8.83 (6849.8)} & \textbf{10.26 (28430.2)} & \textbf{7.74 (2292.0)} \\
 & OWL & 5.59 (267.0) & 6.56 (707.6) & 5.60 (271.5) & 8.71 (6065.7) & 9.66 (15680.2) & 7.72 (2251.0) & 11.60 (109443.9) & 11.24 (75885.4) & 12.43 (250240.0) \\
 & DSA & 5.15 (173.0) & 5.64 (280.5) & 5.17 (176.7) & 8.41 (4500.9) & 9.67 (15803.2) & 7.08 (1185.7) & 10.41 (33179.1) & 10.79 (48424.5) & 10.60 (40099.6) \\
 & LSA & 5.20 (180.6) & 5.86 (349.8) & 5.13 (169.6) & 7.39 (1612.8) & 8.64 (5627.5) & 6.79 (887.2) & 12.06 (173103.3) & 12.27 (213937.1) & 12.08 (176410.1) \\
\midrule
\multicolumn{11}{c}{Dense: WT2=2.54 (12.7), PTB=3.09 (21.9), C4=2.99 (19.9)} \\
\midrule
Llama-3.2-1B & LADDER & \textbf{4.33 (76.2)} & \textbf{4.86 (128.9)} & \textbf{4.37 (79.3)} & \textbf{5.94 (378.1)} & 6.76 (866.4) & \textbf{5.63 (279.0)} & \textbf{8.23 (3740.4)} & 8.81 (6717.7) & \textbf{7.52 (1844.5)} \\
 & OWL & 4.79 (119.8) & 4.99 (146.4) & 4.88 (131.6) & 6.52 (680.5) & 6.67 (789.4) & 5.90 (363.3) & 8.55 (5183.3) & \textbf{8.59 (5397.8)} & 7.74 (2306.7) \\
 & DSA & 5.22 (185.0) & 5.37 (214.4) & 5.08 (161.2) & 6.63 (758.1) & 7.03 (1128.5) & 6.24 (513.4) & 9.50 (13359.7) & 9.62 (15138.6) & 8.60 (5420.0) \\
 & LSA & 4.80 (121.1) & 5.09 (162.7) & 4.95 (141.5) & 6.13 (458.5) & \textbf{6.36 (576.8)} & 6.06 (427.9) & 8.95 (7698.7) & 9.01 (8177.3) & 8.33 (4129.9) \\
\midrule
\multicolumn{11}{c}{Dense: WT2=2.28 (9.8), PTB=2.76 (15.8), C4=2.64 (14.0)} \\
\midrule
OPT-1.3B & LADDER & \textbf{3.82 (45.6)} & 4.36 (78.3) & \textbf{3.60 (36.5)} & \textbf{5.40 (221.8)} & \textbf{5.85 (348.3)} & \textbf{4.74 (114.0)} & \textbf{7.71 (2238.1)} & \textbf{7.74 (2290.5)} & \textbf{6.61 (743.8)} \\
 & OWL & 4.12 (61.7) & 4.33 (75.8) & 3.82 (45.7) & 6.42 (614.5) & 6.21 (495.7) & 5.46 (234.5) & 8.93 (7586.7) & 8.95 (7687.1) & 7.77 (2369.7) \\
 & DSA & 4.10 (60.6) & 4.29 (72.9) & 3.88 (48.3) & 7.44 (1695.1) & 7.44 (1703.7) & 6.22 (500.4) & 8.79 (6572.3) & 8.67 (5799.0) & 7.92 (2752.6) \\
 & LSA & 4.06 (58.2) & \textbf{4.15 (63.6)} & 3.80 (44.6) & 6.45 (629.9) & 6.16 (473.1) & 5.62 (275.1) & 8.78 (6482.0) & 8.42 (4519.0) & 7.81 (2461.7) \\
\midrule
\multicolumn{11}{c}{Dense: WT2=2.69 (14.7), PTB=2.80 (16.5), C4=2.78 (16.1)} \\
\bottomrule
\end{tabular}%
}

\end{table*}
Figure \ref{fig:cross-model-loss-trajectories} presents the pre- and post-distillation trajectories on the same scale.
The distillation process enhances all methods and significantly reduces their differences, particularly at moderate levels of sparsity, while still preserving the general trend observed at higher levels of sparsity.
However, the post-distillation ordering remains architecture-dependent: on Llama-3.2-1B, OWL and LSA are competitive in some high-sparsity settings.
Overall, distillation narrows the differences without eliminating architecture-dependent behavior at high sparsity.
\vspace{-0.5em}
\subsection{Downstream Transfer}
\label{sec:experiments-downstream}
\vspace{-0.5em}
Figure \ref{fig:cross-model-distilled-downstream} presents the complete matched downstream comparison for Qwen3-0.6B, Llama-3.2-1B, and OPT-1.3B.
The shaded regions show the task-level bootstrap standard errors reported by \texttt{lm-eval}.

Across models and tasks, \method{} generally retains accuracy better as sparsity increases, although the strength and ordering of the methods vary across tasks.
When accuracy is macro-averaged across the seven tasks, \method{} generally shows lower degradation from dense performance in the high-sparsity regime. Appendix~\ref{sec:appendix-downstream} reports the complete task-averaged comparison.

Several curves converge near task-specific performance floors, so large language-modeling improvements do not always translate proportionally to downstream accuracy.
We therefore interpret the results as evidence of more consistent downstream retention across the evaluated models and tasks, rather than dominance in every configuration.

\subsection{Ablations}
\label{sec:experiments-ablations}

\begin{figure}[tb]
    \centering
    \includegraphics[width=\columnwidth]{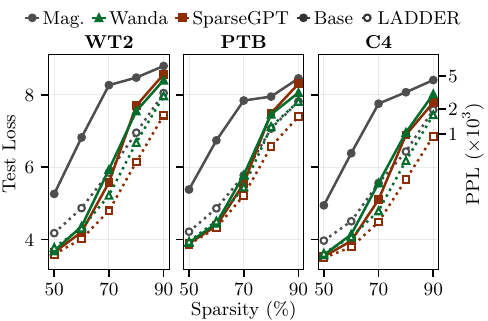}
    \vspace{-2.em}
    \caption{
        Backend ablation on OPT-125M. For each local pruning rule, we compare the standalone method with \method{} allocation.
    }
    \vspace{-1.5em}
    \label{fig:backend-ablation}
\end{figure}

Figure \ref{fig:backend-ablation} shows how \method{} enhances local pruning metrics through its allocation process and identifies the most effective backend rule.
With $\gamma=0.20$, \method{} lowers test loss in 41 of 45 backend--dataset--sparsity comparisons.
It outperforms the corresponding standalone backend across all comparisons at 70\%, 80\%, and 90\% sparsity.
Notably, SparseGPT significantly outperforms other backends, being the reason it was selected for the main experiments.
\begin{figure}[tb]
    \centering
    \includegraphics[width=\columnwidth]{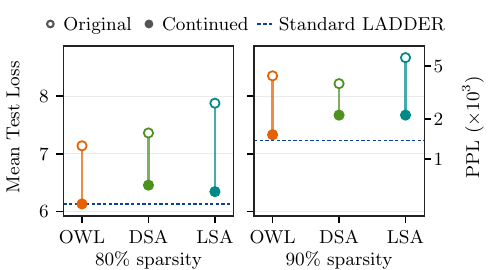}
    \vspace{-1.5em}
    \caption{
        Initializer ablation on OPT-125M, averaged over WikiText-2, PTB, and C4.
        ``Continued'' applies \method{} from each baseline's 70\% mask.
        Dashed lines denote standard \method{}.
    }
    \label{fig:initializer-ablation}
    \vspace{-1.5em}
\end{figure}

Figure \ref{fig:initializer-ablation} shows that \method{} enhances existing non-uniform allocation when loss sharply increases.
Starting from the baselines' 70\% masks, our method achieves lower-loss checkpoints at 80\% and 90\% compared to baseline masks directly generated at those sparsities.
This suggests that a cost-effective allocation can initiate moderate sparsity, with \method{} reserved for challenging high-sparsity adjustments.
Among the tested allocations, OWL offers the strongest initialization, closely aligning with the dense-to-sparse \method{} trajectory at 80\% sparsity.
However, the standard dense-to-sparse \method{} outperforms at 90\%, highlighting how initial allocation decisions can restrict later performance recovery.

We further evaluate the role of the proposed nested trajectory by initializing \method{} for each target sparsity level from the dense checkpoint instead of from the preceding support. 
We observed that the performance of both the independent and progressive variants is comparable at moderate sparsity. However, at the highest sparsity, the nested trajectory reduces perplexity by 10.6\%, 8.6\%, and 15.6\% on WikiText-2, PTB, and C4, respectively.
Importantly, the complete nested trajectory from 10\% to 90\% requires 333 search rounds, approximately the same as the 327 rounds required for a single independently optimized 90\% mask, while simultaneously producing all intermediate sparsity checkpoints.
These results indicate that nesting not only reduces costs relative to an independent search but also provides a better starting point for a high-sparsity search.

\subsection{Scalability and Deployment}
\label{sec:experiments-scope}

\method{} is a one-time post-training procedure that uses a small calibration set without requiring retraining.
Evaluating a single candidate per round makes allocation decisions primarily influenced by depth and tensor granularity.
A naive implementation makes the process costlier with wider layers. Nevertheless, efficient use of hardware allows parallel candidate evaluation.
This post-training search is relatively small compared to pre-training and can be amortized across future deployments, although larger model evaluations are still necessary.
We prioritize pruning quality over direct measurement of inference throughput. At the moment, we rely on existing sparse-kernel results that translate unstructured pruning into accelerated performance \citep{mackoMACKOSparseMatrixvector2026}.

\section{Conclusion}
We introduced \method{}, a post-training allocation process that repeatedly selects the least damaging pruning update as sparsity increases.
This controlled trajectory complements different local pruning rules, with SparseGPT as the strongest backend, while producing nested mask supports across budgets.
Across five models and three language-modeling datasets, our method limits quality degradation in the aggressive 70--90\% sparsity regime compared with OWL, DSA, and LSA.
Distillation narrows these differences, while downstream evaluation shows more consistent accuracy retention across models and tasks.
\method{} can also refine masks initialized by existing allocation methods, although its full dense-to-sparse trajectory remains strongest at 90\%, suggesting the importance of earlier allocation decisions.
These results position our method as a flexible strategy for constructing robust high-sparsity model families.

\subsection*{AI Use Statement}

Generative AI tools were used to assist with software implementation and debugging, edit manuscript prose, and review the manuscript for clarity, internal consistency, and submission compliance.
They were not used to generate experimental measurements or replace the authors' scientific judgment.
The authors reviewed and verified all AI-assisted code and text and take full responsibility for the methods, experiments, claims, and conclusions.

\section*{Reproducibility Statement}

To support reproducibility, an anonymized implementation of LADDER, together with the scripts used to produce all reported results, is available at \url{https://anonymous.4open.science/r/ladder/}.
The repository includes the LADDER allocation procedure (Algorithm 1), the OWL, DSA, and LSA baselines, the posterior-distillation pipeline, and the evaluation code.

\bibliography{refs}
\bibliographystyle{plainnat}

\ifdefined\isarxivpreprint
\else
  \input{sections/08_checklist}
\fi

\clearpage
\onecolumn
\thispagestyle{empty}
\appendix

\section{Supplementary Material}
\label{sec:appendix-experiments}

This appendix provides the experimental settings and the results omitted from the main paper.
It first describes pruning, evaluation, and distillation, then reports the complete language-modeling and downstream results, followed by the ablations and mask analyses.

\subsection{Reproducibility and Exact Configurations}
\label{sec:appendix-reproducibility}

\paragraph{Pruning and allocation configurations.}
Following \citet{frantarSparseGPTMassiveLanguage2023a}, we prune all linear layers within the transformer blocks while excluding biases.
OWL, DSA, LSA, and \method{} use SparseGPT as the common local pruning backend, with 128-column blocks and mean-diagonal Hessian damping of $0.01$.
Local mask selection and weight reconstruction are therefore fixed across these methods, isolating the effect of the global sparsity allocation.

For \method{}, we set the minimum threshold to $z_{\min}=1$ and use a step fraction of $\gamma=0.20$ for OPT and Qwen3-0.6B and $\gamma=0.10$ for Llama-3.2-1B.
For OWL, we set $M=10$, such that its layer-wise outlier distribution counts scores exceeding ten times the corresponding block mean, and use an allocation spread of $\lambda=0.08$.
For DSA, we use the allocation expression
\texttt{W:(ABSLOG)-(VAR)-(ATAN, ASIN)-(7)},
obtained through the original evolutionary search and transferred unchanged to the evaluated models, with $\lambda=0.08$.
For LSA, we use the LSAC variant with $\lambda=0.08$ and construct projection-level reconstruction-error profiles at 50\% probe sparsity.

Table \ref{tab:hyperparameter-development} summarizes the settings shared across the main comparisons, the method-specific configurations, and the configurations used in the ablations.

\begin{table}[t]
    \centering
    \caption{Pruning, evaluation, and ablation configurations.}
    \label{tab:hyperparameter-development}
    \parbox{\textwidth}{%
\small
\setlength{\tabcolsep}{4pt}
\renewcommand{\arraystretch}{1.08}
\newcolumntype{Y}{>{\raggedright\arraybackslash}X}
\begin{tabularx}{\textwidth}{@{}p{0.18\textwidth}Y@{}}
    \toprule
    \multicolumn{2}{@{}l}{\textbf{Shared protocol}} \\
    \midrule
    Models & Pretrained OPT-125M, OPT-350M, OPT-1.3B, Qwen3-0.6B, and Llama-3.2-1B; no additional training before pruning. \\
    Pruning & Unstructured transformer-block \texttt{Linear} weights; biases and output heads excluded; requested sparsity from 10\% to 90\% in 10-point increments. \\
    Calibration and local pruning & 128 C4 training sequences of 2,048 tokens, batch size 1, seed 0; SparseGPT block size 128 and mean-diagonal Hessian damping 0.01. \\
    Evaluation & WikiText-2 and PTB test splits and C4 validation, using contiguous 2,048-token windows; seven zero-shot \texttt{lm-eval} tasks with batch size 1. \\
    Posterior distillation & Dense teacher; logit KL; $T=2$; AdamW with learning rate $10^{-4}$; gradient clipping 1; five epochs over the nine fixed masks. \\
    \bottomrule
\end{tabularx}
\vspace{0.45em}
\begin{tabularx}{\textwidth}{@{}p{0.16\textwidth}YYYY@{}}
    \toprule
    \textbf{Method} & \textbf{\method{}} & \textbf{OWL} & \textbf{DSA} & \textbf{LSA} \\
    \midrule
    Configuration & $\gamma=0.20$ for OPT/Qwen and $0.10$ for Llama; $z_{\min}=1$. & $M=10$ and $\lambda=0.08$. & Expression transferred from the original search; $\lambda=0.08$. & LSAC; $\lambda=0.08$; 50\% probe sparsity; profile block size 128. \\
    Score and granularity & Calibration perplexity for tensor-level pruning choices. & Layer-wise outlier distribution. & Transferred layer-wise analytic function. & Projection-level reconstruction error. \\
    \multicolumn{5}{@{}l}{DSA expression: \texttt{W:(ABSLOG)-(VAR)-(ATAN,ASIN)-(7)}.} \\
    \bottomrule
\end{tabularx}
\vspace{0.45em}
\begin{tabularx}{\textwidth}{@{}p{0.18\textwidth}Y@{}}
    \toprule
    \multicolumn{2}{@{}l}{\textbf{Reported ablations}} \\
    \midrule
    Local pruning rule & OPT-125M with $\gamma=0.20$; Magnitude, Wanda, and SparseGPT are compared alone and within \method{}. \\
    Initialization & OPT-125M OWL, DSA, and LSA masks imported at 70\% sparsity and continued with \method{} to 80\% and 90\%; standard dense-to-sparse \method{} is the reference. \\
    \bottomrule
\end{tabularx}
}%

\end{table}

\paragraph{Calibration dataset.}
All allocation methods use the same 128 sequences of 2,048 tokens sampled from the C4 training set \citep{raffelExploringLimitsTransfer2023}.
These samples are used only to estimate the activation- or curvature-dependent quantities required by the pruning rules.
For \method{}, corpus perplexity on this sample additionally serves as the candidate-selection criterion.
The calibration samples are never used to compute the reported language-modeling loss.
Magnitude and Wanda appear only in the local-pruning-rule ablation.

We pin the first English C4 training shard and load its first 2,048 nonempty documents.
Each document is tokenized with the tokenizer of the evaluated model, followed by the configured space separator, to form candidate 2,048-token windows.
Windows that cross document boundaries are ineligible for calibration.
With seed 0, the sampler repeatedly selects an eligible document and then one of its eligible windows, both with replacement, until it collects 128 examples.
We use batch size 1 and record the sampled document indices and window offsets.
This reproduces the random-document, random-window structure of the reference Wanda and SparseGPT loaders while fixing the exact calibration sample.

\paragraph{Language-modeling evaluation.}
We evaluate corpus-level mean next-token cross-entropy $\mathcal{L}$ and the corresponding perplexity $\exp(\mathcal{L})$ on WikiText-2, PTB, and C4.
Evaluation measures next-token loss on deterministic held-out token streams.
For each split, raw documents are joined using the dataset-specific separator, tokenized once, and divided into consecutive, nonoverlapping 2,048-token windows; the incomplete tail is discarded.
Thus, every pruning method is scored on the same target tokens:

\begin{itemize}
    \item \textbf{WikiText-2 (WT2):} the \texttt{Salesforce/wikitext} dataset with configuration \texttt{wikitext-2-raw-v1}; we score the official test split.
    \item \textbf{Penn Treebank (PTB):} the public Zaremba text files; we score the official test split.
    \item \textbf{C4:} the English \texttt{allenai/c4} corpus. We pin the first validation shard, join its first 1,100 documents, and retain the first 256 complete windows. Calibration uses the separate training shard described above.
\end{itemize}

\paragraph{Posterior-distillation objective.}
Posterior distillation fixes the nine masks corresponding to sparsities from 10\% to 90\% and jointly optimizes a single shared student state.
The student is initialized from the pretrained dense weights; method-specific SparseGPT reconstructed weights are not used for initialization.

Let $\Scand=\{s_1,\ldots,s_J\}$ denote the participating sparsity budgets and let $p_t^{(T)}$ and $p_s^{(T)}$ be the dense-teacher and masked-student distributions at temperature $T$.
The shared student minimizes
\begin{align*}
    \textstyle
    \Lcal_{\mathrm{KD}}
    = \frac{T^2}{|\Scand|}\sum_{s\in\Scand}
    \KL\!\left(p_t^{(T)}(\cdot\mid x)\,\|\,p_s^{(T)}(\cdot\mid x)\right).
\end{align*}
Every participating mask is evaluated against the same teacher outputs, the mask-specific losses are averaged, and one optimizer step updates the shared parameters.
We train for five epochs using AdamW with a learning rate of $10^{-4}$ and temperature $T=2$.

\paragraph{Implementation details.}
All pruning and distillation runs use seed 0 with deterministic algorithms, so each reported model--method pair represents one run rather than an average over independent runs.
Distillation reuses the 128 calibration sequences; for each batch, equally weighted gradients are accumulated over the nine fixed masks before one AdamW update, giving 128 updates per epoch and 640 updates in total.
We use gradient checkpointing.

\paragraph{Computing infrastructure.}
Experiments were run on compute nodes equipped with four NVIDIA A100-SXM4 GPUs with 40~GB of memory each, using NVIDIA driver 570.195.03 and CUDA 12.8.

\subsection{Downstream Summary}
\label{sec:appendix-downstream}

We use \texttt{lm-eval} 0.4.12 with zero demonstrations and batch size 1 on seven tasks:

\begin{itemize}
    \item \textbf{ARC-Easy and ARC-Challenge:} grade-school multiple-choice science questions in easier and reasoning-intensive subsets.
    \item \textbf{BoolQ:} binary reading comprehension that asks whether a passage supports a question.
    \item \textbf{HellaSwag:} commonsense completion by selecting the most plausible continuation.
    \item \textbf{OpenBookQA:} multiple-choice elementary science questions based on a small collection of core facts.
    \item \textbf{PIQA:} physical commonsense reasoning that selects the more plausible solution to a practical goal.
    \item \textbf{WinoGrande:} commonsense coreference resolution in ambiguity-controlled sentence pairs.
\end{itemize}

BoolQ uses task version 1; the other tasks use version 0.
We report normalized accuracy for the multiple-choice tasks and ordinary accuracy for BoolQ and WinoGrande.
For aggregate reporting, we take the unweighted mean of the seven task accuracies and combine their task-level bootstrap standard errors.
Repeated evaluation seeds for the same model state produced identical accuracies. The uncertainty shown in the figures is therefore the task-level bootstrap standard error returned by \texttt{lm-eval}, not variation across independently pruned models.
\begin{figure}[t]
    \centering
    \includegraphics[width=\linewidth]{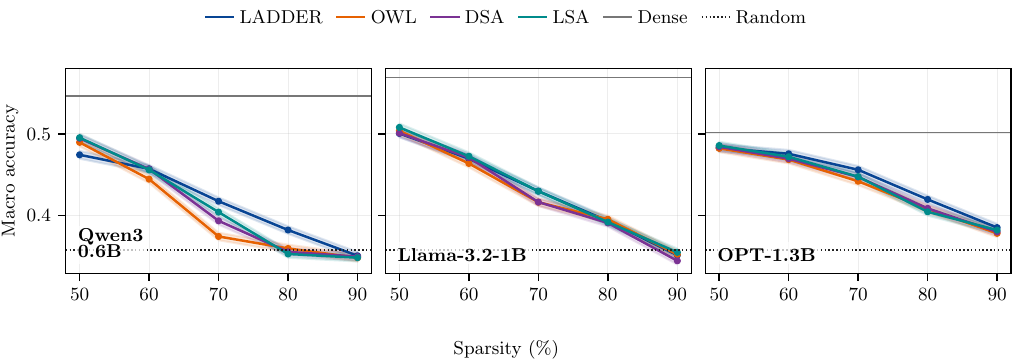}
    \caption{
        Mean post-distillation zero-shot accuracy over seven tasks for Qwen3-0.6B, Llama-3.2-1B, and OPT-1.3B.
        Curves compare methods from 50\% to 90\% sparsity; solid horizontal lines mark dense accuracy, while the dotted line marks the seven-task mean random accuracy (35.7\%).
        Shading denotes propagated \texttt{lm-eval} uncertainty.
    }
    \label{fig:appendix-cross-model-distilled-downstream-average}
\end{figure}

In Figure \ref{fig:appendix-cross-model-distilled-downstream-average}, \method{} has the highest mean accuracy in seven of the nine model--sparsity comparisons at 70--90\%.
OWL is highest for Llama-3.2-1B at 80\%, and LSA is highest for that model at 90\%.

\subsection{Ablations}
\label{sec:appendix-ablations}

\paragraph{Local pruning rule and external initialization.}
Table \ref{tab:backend-ablation-baseline-comparison-table} compares each local pruning rule used alone and within \method{}.
At 70--90\% sparsity, \method{} lowers test loss in all 27 matched rule--dataset comparisons, and SparseGPT gives the lowest \method{} loss in all nine dataset--sparsity comparisons.

Table \ref{tab:initializer-ablation-loss-table} starts \method{} from the 70\% masks produced by OWL, DSA, or LSA.
Continuing with \method{} lowers loss relative to extending the original method in all 18 comparisons.
OWL gives the strongest external starting point, while the dense-to-sparse \method{} trajectory gives the lowest loss on every dataset at 90\%.

\begin{table*}[t]
    \caption{
        Each local pruning rule used alone (Base) and within LADDER. Values are test loss; bold marks the better matched result.
    }
    \label{tab:backend-ablation-baseline-comparison-table}
    \begin{center}
    \small
    \begin{tabular}{llrrrrrr}
        \toprule
        Sparsity & Local rule & \multicolumn{3}{c}{Base} & \multicolumn{3}{c}{LADDER} \\
        \cmidrule(lr){3-5} \cmidrule(lr){6-8}
        &  & WT2 & PTB & C4 & WT2 & PTB & C4 \\
        \midrule
        50\% & Mag. & 5.26 & 5.38 & 4.94 & \textbf{4.18} & \textbf{4.22} & \textbf{3.97} \\
        & SparseGPT & 3.65 & 3.89 & 3.53 & \textbf{3.59} & \textbf{3.87} & \textbf{3.50} \\
        & Wanda & \textbf{3.69} & \textbf{3.92} & \textbf{3.58} & 3.78 & 3.95 & 3.62 \\
        \midrule
        60\% & Mag. & 6.82 & 6.74 & 6.38 & \textbf{4.87} & \textbf{4.86} & \textbf{4.50} \\
        & SparseGPT & 4.21 & 4.38 & 3.96 & \textbf{4.02} & \textbf{4.33} & \textbf{3.79} \\
        & Wanda & 4.36 & \textbf{4.45} & 4.14 & \textbf{4.31} & 4.51 & \textbf{4.07} \\
        \midrule
        70\% & Mag. & 8.26 & 7.84 & 7.75 & \textbf{5.85} & \textbf{5.77} & \textbf{5.56} \\
        & SparseGPT & 5.58 & 5.55 & 5.10 & \textbf{4.79} & \textbf{5.22} & \textbf{4.48} \\
        & Wanda & 5.93 & 5.78 & 5.56 & \textbf{5.22} & \textbf{5.45} & \textbf{4.79} \\
        \midrule
        80\% & Mag. & 8.47 & 7.95 & 8.07 & \textbf{6.95} & \textbf{7.07} & \textbf{6.43} \\
        & SparseGPT & 7.70 & 7.49 & 6.89 & \textbf{6.14} & \textbf{6.57} & \textbf{5.66} \\
        & Wanda & 7.55 & 7.46 & 6.96 & \textbf{6.68} & \textbf{7.12} & \textbf{6.18} \\
        \midrule
        90\% & Mag. & 8.79 & 8.45 & 8.40 & \textbf{8.05} & \textbf{7.81} & \textbf{7.58} \\
        & SparseGPT & 8.56 & 8.30 & 7.78 & \textbf{7.43} & \textbf{7.40} & \textbf{6.85} \\
        & Wanda & 8.40 & 8.06 & 8.02 & \textbf{7.97} & \textbf{7.81} & \textbf{7.44} \\
        \midrule
        \multicolumn{8}{c}{Dense: WT2=3.32, PTB=3.41, C4=3.28.} \\
        \bottomrule
    \end{tabular}
    \end{center}
\end{table*}

\begin{table*}[t]
    \caption{
        Dataset-level test loss for direct baselines, LADDER continuation from their 70\% masks, and the standard dense-to-sparse LADDER trajectory.
    }
    \label{tab:initializer-ablation-loss-table}
    \begin{center}
    \small
    \begin{tabular}{lrrrrrr}
        \toprule
        Configuration & \multicolumn{3}{c}{80\%} & \multicolumn{3}{c}{90\%} \\
        \cmidrule(lr){2-4} \cmidrule(lr){5-7}
        & WT2 & PTB & C4 & WT2 & PTB & C4 \\
        \midrule
        Standard LADDER & \textbf{6.14} & 6.57 & 5.66 & \textbf{7.43} & \textbf{7.40} & \textbf{6.85} \\
        OWL (direct) & 7.48 & 7.30 & 6.64 & 8.68 & 8.31 & 8.06 \\
        OWL $\rightarrow$ LADDER & 6.33 & \textbf{6.41} & \textbf{5.65} & 7.57 & 7.52 & 6.91 \\
        DSA (direct) & 7.70 & 7.49 & 6.89 & 8.56 & 8.30 & 7.78 \\
        DSA $\rightarrow$ LADDER & 6.60 & 6.88 & 5.89 & 7.86 & 7.90 & 7.25 \\
        LSA (direct) & 8.24 & 7.92 & 7.47 & 8.92 & 8.62 & 8.47 \\
        LSA $\rightarrow$ LADDER & 6.53 & 6.75 & 5.74 & 7.91 & 7.87 & 7.23 \\
        \midrule
        \multicolumn{7}{c}{Dense: WT2=3.32, PTB=3.41, C4=3.28.} \\
        \bottomrule
    \end{tabular}
    \end{center}
\end{table*}

\subsection{Complete Language-Modeling Results}
\label{sec:appendix-loss}

\begin{figure*}[t]
    \centering
    \begin{minipage}[t]{0.48\textwidth}\centering
        \includegraphics[width=\linewidth]{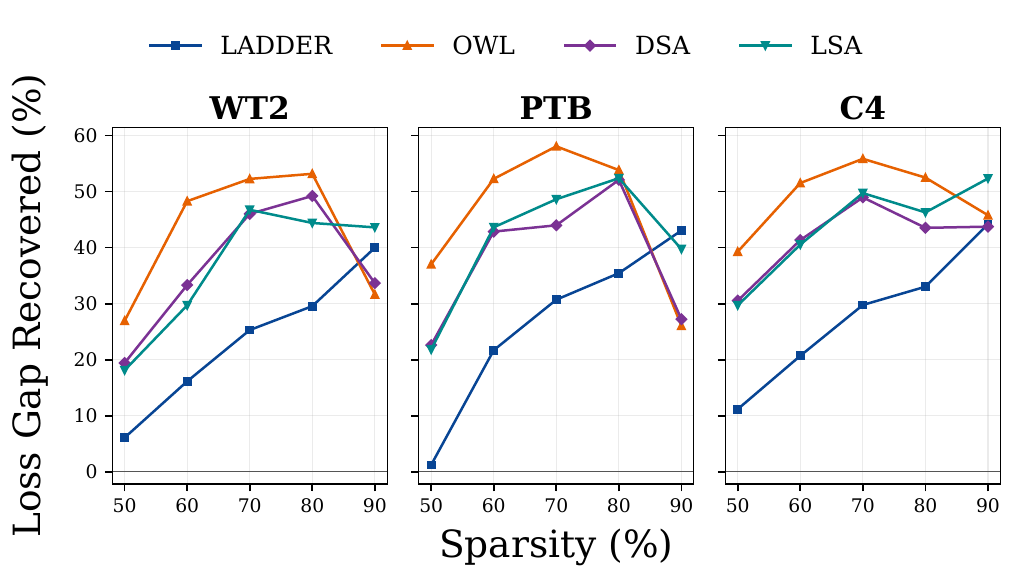}
        \par\smallskip (a) Qwen3-0.6B
    \end{minipage}\hfill
    \begin{minipage}[t]{0.48\textwidth}\centering
        \includegraphics[width=\linewidth]{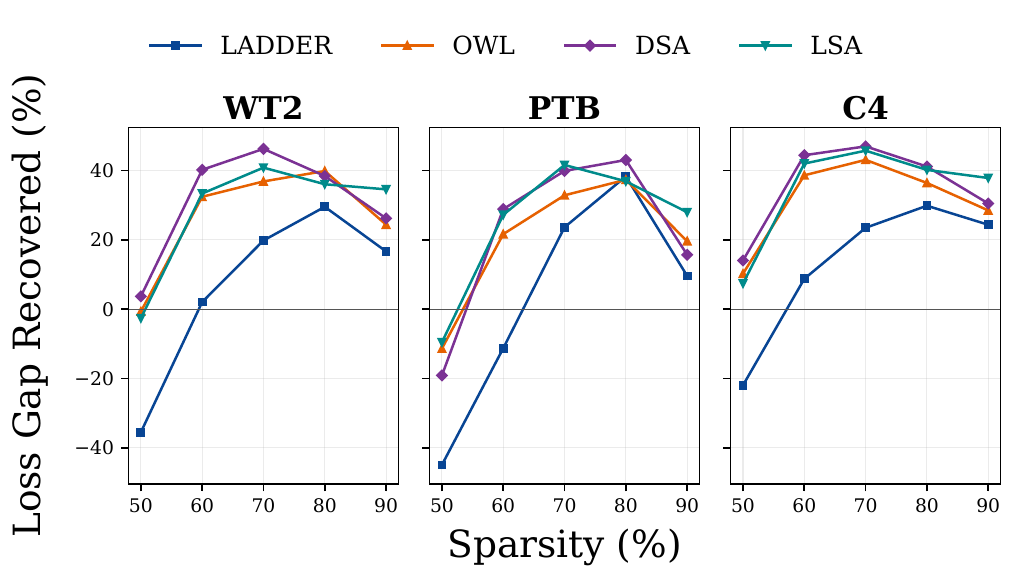}
        \par\smallskip (b) Llama-3.2-1B
    \end{minipage}\\[-0.1em]
    \begin{minipage}[t]{0.48\textwidth}\centering
        \includegraphics[width=\linewidth]{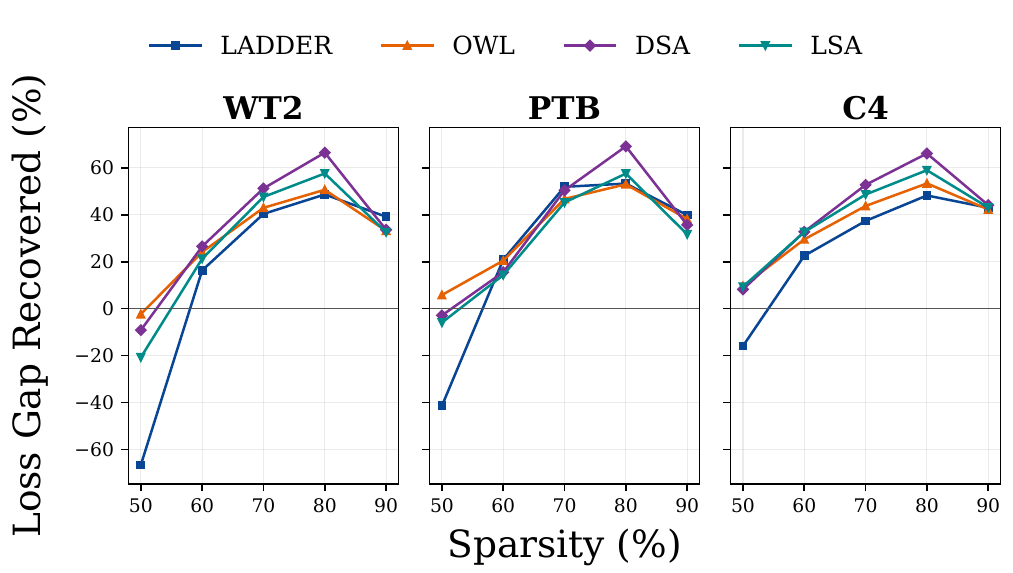}
        \par\smallskip (c) OPT-1.3B
    \end{minipage}\hfill
    \begin{minipage}[t]{0.48\textwidth}\centering
        \includegraphics[width=\linewidth]{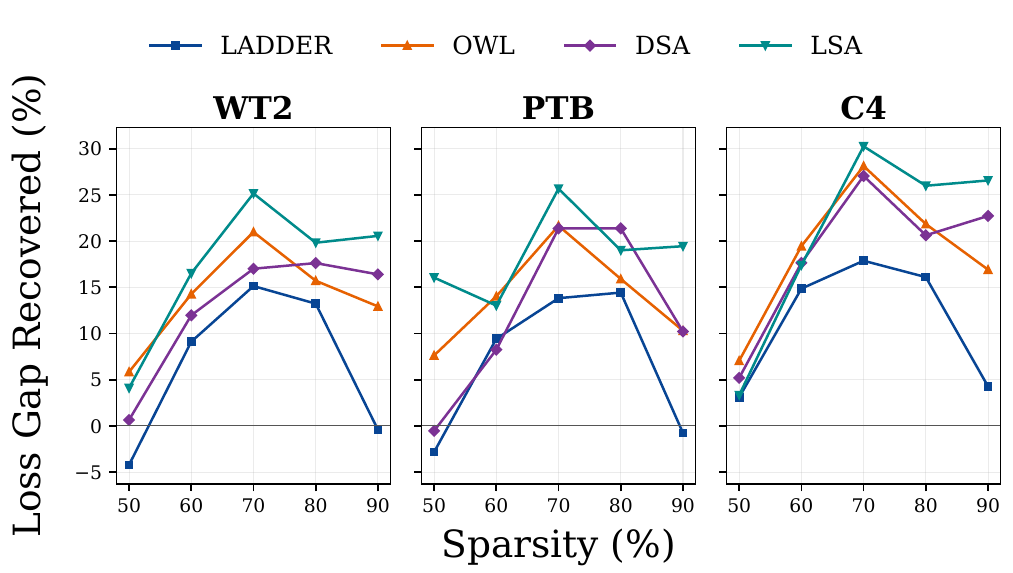}
        \par\smallskip (d) OPT-350M
    \end{minipage}\\[-0.1em]
    \begin{minipage}[t]{0.48\textwidth}\centering
        \includegraphics[width=\linewidth]{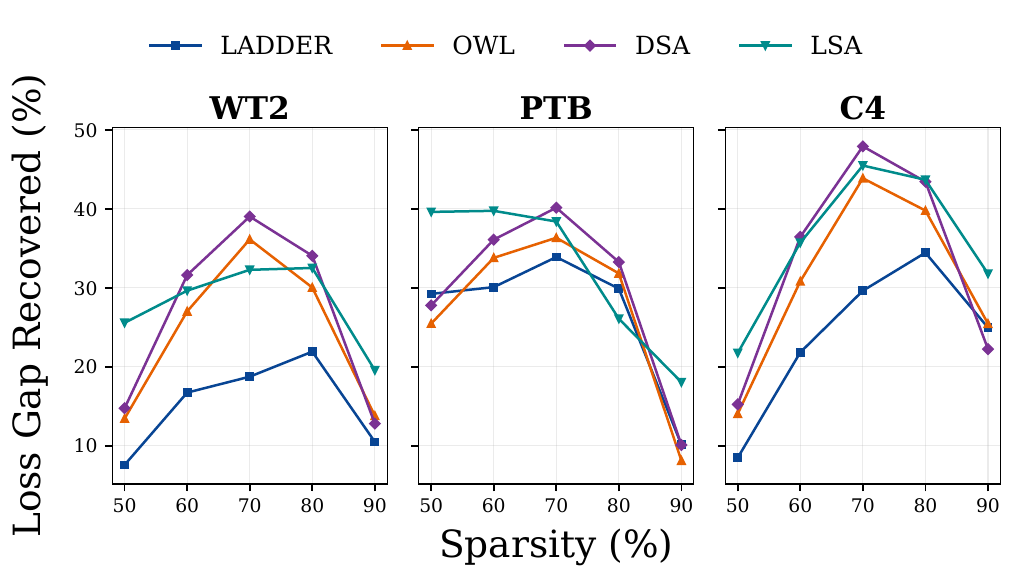}
        \par\smallskip (e) OPT-125M
    \end{minipage}
    \caption{
        Fraction of the pruning-induced test-loss gap removed by posterior distillation, $100(L_{\mathrm{pre}}-L_{\mathrm{post}})/(L_{\mathrm{pre}}-L_{\mathrm{dense}})$.
        Higher is better; zero indicates no recovery.
        }
    \label{fig:appendix-distillation-gap-recovery}
\end{figure*}

Across the 180 model, dataset, method, and sparsity combinations at 70--90\%, posterior distillation reduces test loss in 178 cases.
The amount recovered varies substantially, and the two exceptions are both 90\% sparse OPT-350M \method{} models.

Table \ref{tab:cross-model-pruned-lm-performance-full} reports the pre-distillation results for all five evaluated models.

\begin{table*}[t]
    \center
    \caption{
        Complete pre-distillation language-modeling results at 70--90\% sparsity.
        Entries report test loss followed by perplexity in parentheses; lower is better, and bold marks the best method for each configuration.
    }
    \label{tab:cross-model-pruned-lm-performance-full}
    \resizebox{\textwidth}{!}{%
\begin{tabular}{llccccccccc}
\toprule
Model & Method & \multicolumn{3}{c}{70\%} & \multicolumn{3}{c}{80\%} & \multicolumn{3}{c}{90\%} \\
\cmidrule(lr){3-5} \cmidrule(lr){6-8} \cmidrule(lr){9-11}
 & & WT2 & PTB & C4 & WT2 & PTB & C4 & WT2 & PTB & C4 \\
\midrule
OPT-125M & LADDER & \textbf{4.79 (120.6)} & \textbf{5.22 (184.6)} & \textbf{4.48 (88.5)} & \textbf{6.14 (465.4)} & \textbf{6.57 (712.8)} & \textbf{5.66 (286.6)} & \textbf{7.43 (1693.2)} & \textbf{7.40 (1629.2)} & \textbf{6.85 (947.3)} \\
 & OWL & 5.54 (254.3) & 5.49 (243.3) & 4.98 (145.2) & 7.48 (1772.1) & 7.30 (1473.4) & 6.64 (764.1) & 8.68 (5862.5) & 8.31 (4082.0) & 8.06 (3179.8) \\
 & DSA & 5.58 (264.2) & 5.55 (256.5) & 5.10 (164.4) & 7.70 (2205.9) & 7.49 (1793.7) & 6.89 (983.1) & 8.56 (5236.4) & 8.30 (4029.8) & 7.78 (2403.5) \\
 & LSA & 6.37 (581.6) & 6.58 (718.1) & 5.58 (264.7) & 8.24 (3798.7) & 7.92 (2739.3) & 7.47 (1760.6) & 8.92 (7447.4) & 8.62 (5535.1) & 8.47 (4765.3) \\
\midrule
\multicolumn{11}{c}{Dense: WT2=3.32 (27.7), PTB=3.41 (30.2), C4=3.28 (26.6)} \\
\midrule
OPT-350M & LADDER & \textbf{4.61 (100.2)} & \textbf{4.51 (91.0)} & \textbf{4.27 (71.7)} & \textbf{5.63 (279.8)} & \textbf{5.27 (194.7)} & \textbf{5.21 (182.7)} & \textbf{7.20 (1340.0)} & \textbf{6.57 (716.8)} & \textbf{6.48 (651.7)} \\
 & OWL & 5.08 (161.4) & 5.09 (162.2) & 4.70 (109.9) & 6.96 (1057.2) & 6.79 (885.5) & 6.08 (437.2) & 8.58 (5330.8) & 8.19 (3600.1) & 7.63 (2062.6) \\
 & DSA & 5.00 (148.2) & 4.99 (146.2) & 4.68 (107.5) & 7.07 (1181.0) & 6.94 (1029.6) & 6.12 (456.6) & 8.91 (7384.5) & 8.40 (4439.4) & 8.01 (3023.1) \\
 & LSA & 5.11 (166.3) & 4.97 (143.5) & 4.79 (120.3) & 7.16 (1289.6) & 6.87 (961.3) & 6.39 (593.4) & 9.22 (10084.5) & 8.57 (5257.5) & 8.49 (4866.4) \\
\midrule
\multicolumn{11}{c}{Dense: WT2=3.09 (22.0), PTB=3.20 (24.5), C4=3.12 (22.6)} \\
\midrule
OPT-1.3B & LADDER & \textbf{3.82 (45.6)} & 4.36 (78.3) & \textbf{3.60 (36.5)} & \textbf{5.40 (221.8)} & \textbf{5.85 (348.3)} & \textbf{4.74 (114.0)} & \textbf{7.71 (2238.1)} & \textbf{7.74 (2290.5)} & \textbf{6.61 (743.8)} \\
 & OWL & 4.12 (61.7) & 4.33 (75.8) & 3.82 (45.7) & 6.42 (614.5) & 6.21 (495.7) & 5.46 (234.5) & 8.93 (7586.7) & 8.95 (7687.1) & 7.77 (2369.7) \\
 & DSA & 4.10 (60.6) & 4.29 (72.9) & 3.88 (48.3) & 7.44 (1695.1) & 7.44 (1703.7) & 6.22 (500.4) & 8.79 (6572.3) & 8.67 (5799.0) & 7.92 (2752.6) \\
 & LSA & 4.06 (58.2) & \textbf{4.15 (63.6)} & 3.80 (44.6) & 6.45 (629.9) & 6.16 (473.1) & 5.62 (275.1) & 8.78 (6482.0) & 8.42 (4519.0) & 7.81 (2461.7) \\
\midrule
\multicolumn{11}{c}{Dense: WT2=2.69 (14.7), PTB=2.80 (16.5), C4=2.78 (16.1)} \\
\midrule
Qwen3-0.6B & LADDER & \textbf{4.34 (76.4)} & \textbf{5.00 (147.7)} & \textbf{4.40 (81.8)} & \textbf{5.34 (208.7)} & \textbf{6.26 (524.5)} & \textbf{5.29 (197.6)} & \textbf{8.83 (6849.8)} & \textbf{10.26 (28430.2)} & \textbf{7.74 (2292.0)} \\
 & OWL & 5.59 (267.0) & 6.56 (707.6) & 5.60 (271.5) & 8.71 (6065.7) & 9.66 (15680.2) & 7.72 (2251.0) & 11.60 (109443.9) & 11.24 (75885.4) & 12.43 (250240.0) \\
 & DSA & 5.15 (173.0) & 5.64 (280.5) & 5.17 (176.7) & 8.41 (4500.9) & 9.67 (15803.2) & 7.08 (1185.7) & 10.41 (33179.1) & 10.79 (48424.5) & 10.60 (40099.6) \\
 & LSA & 5.20 (180.6) & 5.86 (349.8) & 5.13 (169.6) & 7.39 (1612.8) & 8.64 (5627.5) & 6.79 (887.2) & 12.06 (173103.3) & 12.27 (213937.1) & 12.08 (176410.1) \\
\midrule
\multicolumn{11}{c}{Dense: WT2=2.54 (12.7), PTB=3.09 (21.9), C4=2.99 (19.9)} \\
\midrule
Llama-3.2-1B & LADDER & \textbf{4.33 (76.2)} & \textbf{4.86 (128.9)} & \textbf{4.37 (79.3)} & \textbf{5.94 (378.1)} & 6.76 (866.4) & \textbf{5.63 (279.0)} & \textbf{8.23 (3740.4)} & 8.81 (6717.7) & \textbf{7.52 (1844.5)} \\
 & OWL & 4.79 (119.8) & 4.99 (146.4) & 4.88 (131.6) & 6.52 (680.5) & 6.67 (789.4) & 5.90 (363.3) & 8.55 (5183.3) & \textbf{8.59 (5397.8)} & 7.74 (2306.7) \\
 & DSA & 5.22 (185.0) & 5.37 (214.4) & 5.08 (161.2) & 6.63 (758.1) & 7.03 (1128.5) & 6.24 (513.4) & 9.50 (13359.7) & 9.62 (15138.6) & 8.60 (5420.0) \\
 & LSA & 4.80 (121.1) & 5.09 (162.7) & 4.95 (141.5) & 6.13 (458.5) & \textbf{6.36 (576.8)} & 6.06 (427.9) & 8.95 (7698.7) & 9.01 (8177.3) & 8.33 (4129.9) \\
\midrule
\multicolumn{11}{c}{Dense: WT2=2.28 (9.8), PTB=2.76 (15.8), C4=2.64 (14.0)} \\
\bottomrule
\end{tabular}%
}

\end{table*}

Table \ref{tab:cross-model-distilled-lm-performance-full} reports the corresponding results after posterior distillation.

\begin{table*}[t]
    \center
    \caption{
        Complete post-distillation language-modeling results at 70--90\% sparsity.
        Entries report test loss followed by perplexity in parentheses; lower is better, and bold marks the best method for each configuration.
        }
    \label{tab:cross-model-distilled-lm-performance-full}
    \resizebox{\textwidth}{!}{%
\begin{tabular}{llccccccccc}
\toprule
Model & Method & \multicolumn{3}{c}{70\%} & \multicolumn{3}{c}{80\%} & \multicolumn{3}{c}{90\%} \\
\cmidrule(lr){3-5} \cmidrule(lr){6-8} \cmidrule(lr){9-11}
 & & WT2 & PTB & C4 & WT2 & PTB & C4 & WT2 & PTB & C4 \\
\midrule
OPT-125M & LADDER & \textbf{4.52 (91.5)} & \textbf{4.60 (100.0)} & \textbf{4.13 (61.9)} & \textbf{5.52 (250.6)} & \textbf{5.62 (277.1)} & \textbf{4.84 (126.3)} & \textbf{7.00 (1099.1)} & \textbf{6.99 (1087.3)} & \textbf{5.96 (388.2)} \\
 & OWL & 4.74 (114.0) & 4.74 (114.0) & 4.23 (68.9) & 6.23 (507.8) & 6.06 (427.6) & 5.30 (200.8) & 7.94 (2798.5) & 7.92 (2740.9) & 6.85 (939.9) \\
 & DSA & 4.70 (109.5) & 4.69 (108.7) & 4.23 (68.6) & 6.21 (496.3) & 6.13 (461.1) & 5.32 (204.8) & 7.89 (2674.5) & 7.81 (2457.7) & 6.78 (882.0) \\
 & LSA & 5.38 (217.5) & 5.36 (212.8) & 4.53 (93.0) & 6.64 (766.3) & 6.74 (845.2) & 5.64 (281.8) & 7.82 (2491.2) & 7.68 (2161.0) & 6.82 (914.9) \\
\midrule
\multicolumn{11}{c}{Dense: WT2=3.32 (27.7), PTB=3.41 (30.2), C4=3.28 (26.6)} \\
\midrule
OPT-350M & LADDER & \textbf{4.38 (79.7)} & \textbf{4.33 (75.9)} & \textbf{4.07 (58.3)} & \textbf{5.30 (199.8)} & \textbf{4.97 (144.4)} & \textbf{4.87 (130.5)} & \textbf{7.22 (1362.7)} & \textbf{6.60 (735.5)} & \textbf{6.34 (564.5)} \\
 & OWL & 4.67 (106.3) & 4.68 (107.7) & 4.25 (70.4) & 6.36 (575.4) & 6.22 (500.5) & 5.43 (228.9) & 7.87 (2620.7) & 7.67 (2153.0) & 6.87 (961.3) \\
 & DSA & 4.67 (107.1) & 4.60 (99.8) & 4.26 (70.5) & 6.37 (585.4) & 6.14 (463.1) & 5.50 (245.6) & 7.95 (2845.1) & 7.87 (2609.8) & 6.90 (993.6) \\
 & LSA & 4.61 (100.0) & 4.51 (91.2) & 4.28 (72.6) & 6.36 (575.4) & 6.17 (478.6) & 5.54 (253.8) & 7.96 (2860.4) & 7.52 (1850.6) & 7.06 (1167.9) \\
\midrule
\multicolumn{11}{c}{Dense: WT2=3.09 (22.0), PTB=3.20 (24.5), C4=3.12 (22.6)} \\
\midrule
OPT-1.3B & LADDER & \textbf{3.36 (28.9)} & 3.55 (34.9) & \textbf{3.29 (26.9)} & \textbf{4.08 (59.2)} & \textbf{4.23 (68.5)} & \textbf{3.79 (44.4)} & \textbf{5.74 (312.5)} & \textbf{5.77 (320.4)} & \textbf{4.96 (143.2)} \\
 & OWL & 3.51 (33.3) & 3.61 (37.1) & 3.37 (29.0) & 4.53 (92.7) & 4.40 (81.7) & 4.03 (56.2) & 6.86 (953.8) & 6.60 (735.5) & 5.66 (287.4) \\
 & DSA & 3.38 (29.3) & \textbf{3.54 (34.4)} & 3.30 (27.1) & 4.28 (72.1) & 4.23 (68.9) & 3.94 (51.6) & 6.74 (844.6) & 6.57 (716.8) & 5.65 (285.2) \\
 & LSA & 3.41 (30.2) & 3.54 (34.5) & 3.30 (27.2) & 4.28 (72.2) & 4.23 (68.5) & 3.94 (51.6) & 6.80 (893.4) & 6.64 (763.0) & 5.64 (282.1) \\
\midrule
\multicolumn{11}{c}{Dense: WT2=2.69 (14.7), PTB=2.80 (16.5), C4=2.78 (16.1)} \\
\midrule
Qwen3-0.6B & LADDER & \textbf{3.88 (48.5)} & \textbf{4.41 (82.2)} & \textbf{3.98 (53.7)} & \textbf{4.51 (91.2)} & \textbf{5.14 (170.2)} & \textbf{4.53 (92.6)} & \textbf{6.32 (553.2)} & \textbf{7.17 (1298.1)} & \textbf{5.64 (281.3)} \\
 & OWL & 3.99 (54.3) & 4.54 (94.0) & 4.14 (63.1) & 5.43 (227.9) & 6.12 (455.4) & 5.24 (188.1) & 8.74 (6216.8) & 9.11 (9075.0) & 8.11 (3321.4) \\
 & DSA & 3.95 (51.9) & 4.51 (91.4) & 4.11 (60.7) & 5.52 (249.9) & 6.24 (512.6) & 5.30 (200.1) & 7.76 (2344.5) & 8.69 (5943.8) & 7.27 (1437.6) \\
 & LSA & 3.95 (52.1) & 4.51 (90.9) & 4.07 (58.4) & 5.23 (187.5) & 5.73 (307.5) & 5.03 (153.0) & 7.91 (2719.4) & 8.62 (5554.7) & 7.32 (1514.4) \\
\midrule
\multicolumn{11}{c}{Dense: WT2=2.54 (12.7), PTB=3.09 (21.9), C4=2.99 (19.9)} \\
\midrule
Llama-3.2-1B & LADDER & 3.93 (50.7) & 4.36 (78.5) & 3.97 (52.8) & 4.86 (128.4) & 5.23 (186.9) & 4.74 (114.3) & 7.24 (1389.5) & 8.23 (3760.9) & 6.33 (562.1) \\
 & OWL & 3.86 (47.6) & 4.25 (70.4) & 3.91 (50.1) & 4.83 (125.4) & 5.21 (183.1) & 4.71 (111.1) & 7.02 (1123.4) & 7.45 (1720.8) & 6.29 (539.1) \\
 & DSA & 3.86 (47.5) & 4.33 (75.7) & 3.94 (51.2) & 4.96 (142.4) & 5.19 (179.6) & 4.76 (116.7) & 7.61 (2015.0) & 8.55 (5157.3) & 6.78 (881.5) \\
 & LSA & \textbf{3.77 (43.3)} & \textbf{4.12 (61.6)} & \textbf{3.89 (49.1)} & \textbf{4.74 (114.5)} & \textbf{5.03 (152.9)} & \textbf{4.69 (108.3)} & \textbf{6.64 (767.5)} & \textbf{7.26 (1423.8)} & \textbf{6.17 (480.3)} \\
\midrule
\multicolumn{11}{c}{Dense: WT2=2.28 (9.8), PTB=2.76 (15.8), C4=2.64 (14.0)} \\
\bottomrule
\end{tabular}%
}

\end{table*}

\subsection{Task-Level Downstream Results}

\begin{figure*}[t]
    \centering
    \includegraphics[width=0.98\textwidth]{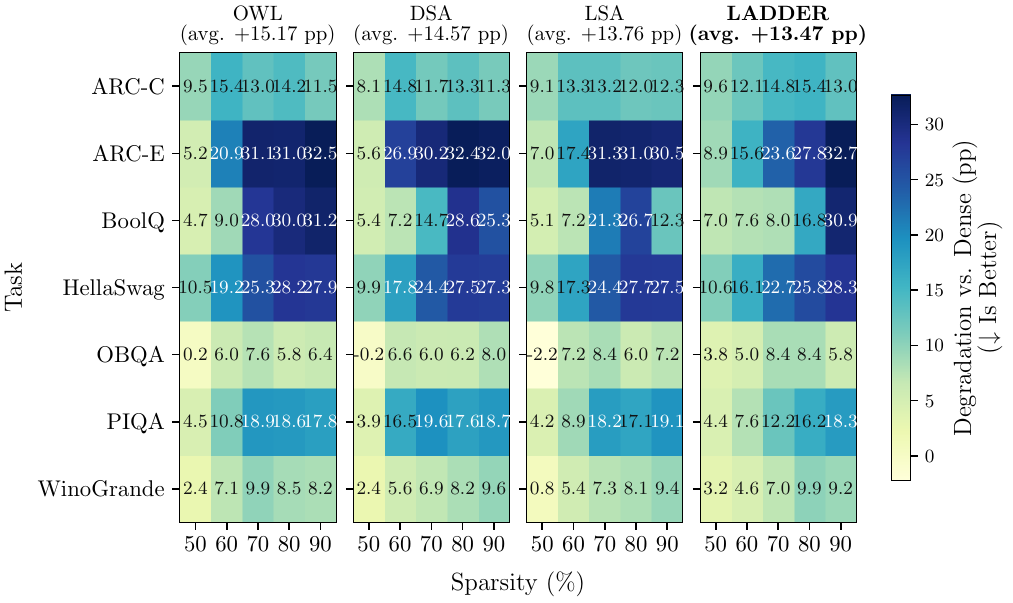}\\[-0.2em]
    {\small (a) Before posterior distillation}\\[0.25em]
    \includegraphics[width=0.98\textwidth]{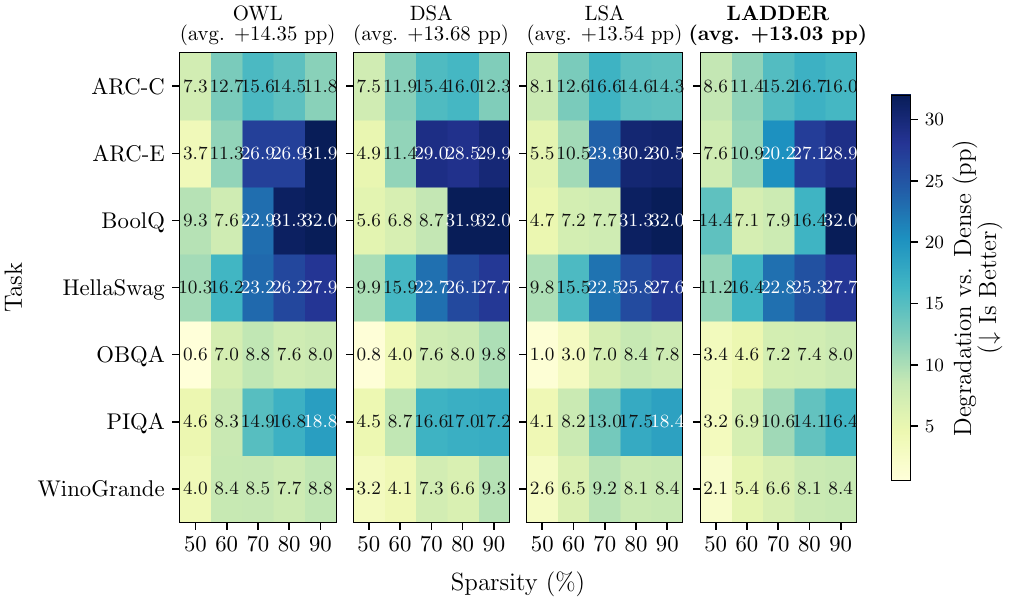}\\[-0.2em]
    {\small (b) After posterior distillation}
    \caption{
        Qwen3-0.6B zero-shot accuracy degradation before and after posterior distillation.
        Lower is better.
    }    
    \label{fig:appendix-qwen-downstream-degradation}
\end{figure*}

\begin{figure*}[t]
    \centering
    \includegraphics[width=0.98\textwidth]{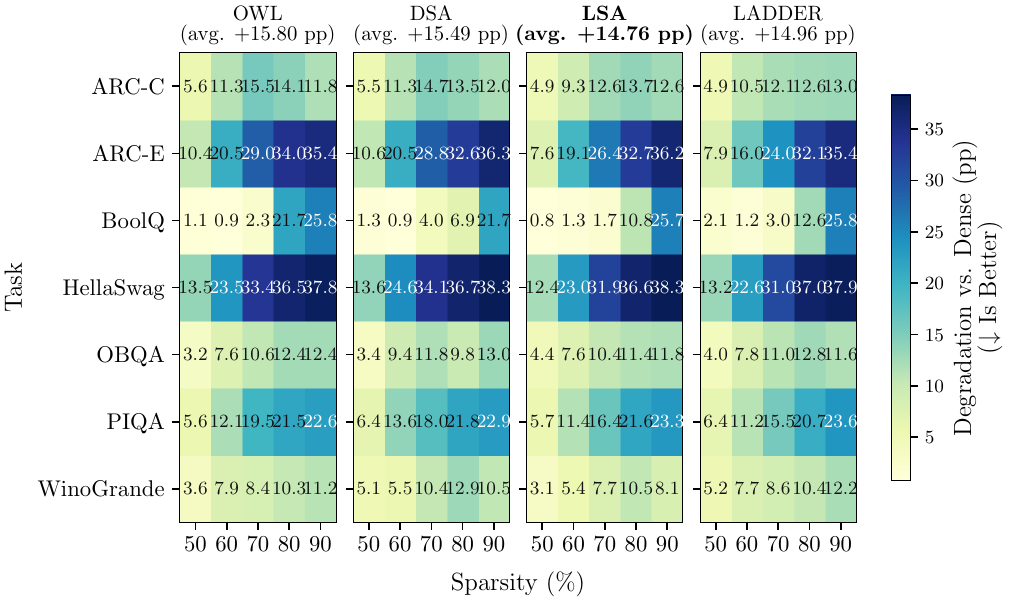}\\[-0.2em]
    {\small (a) Before posterior distillation}\\[0.25em]
    \includegraphics[width=0.98\textwidth]{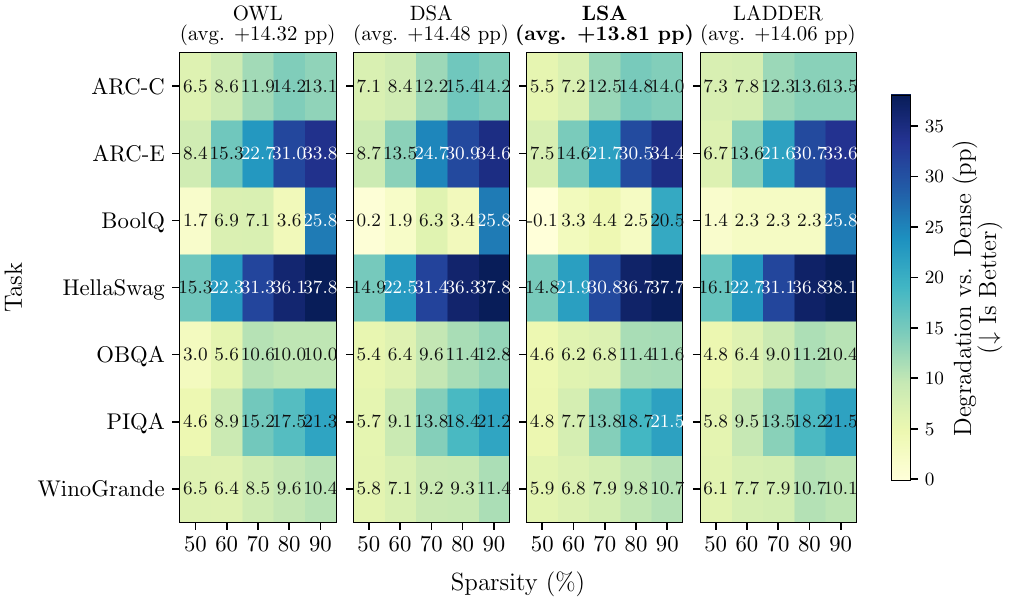}\\[-0.2em]
    {\small (b) After posterior distillation}
    \caption{
        Llama-3.2-1B zero-shot accuracy degradation before and after posterior distillation.
        Lower is better.
    }
    \label{fig:appendix-llama-downstream-degradation}
\end{figure*}

\begin{figure*}[t]
    \centering
    \includegraphics[width=0.98\textwidth]{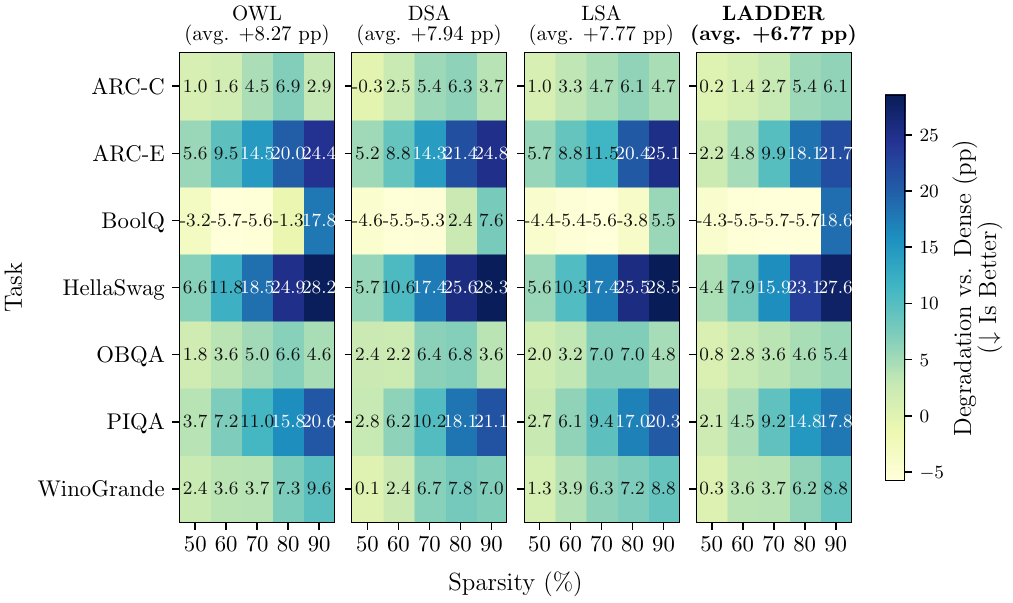}\\[-0.2em]
    {\small (a) Before posterior distillation}\\[0.25em]
    \includegraphics[width=0.98\textwidth]{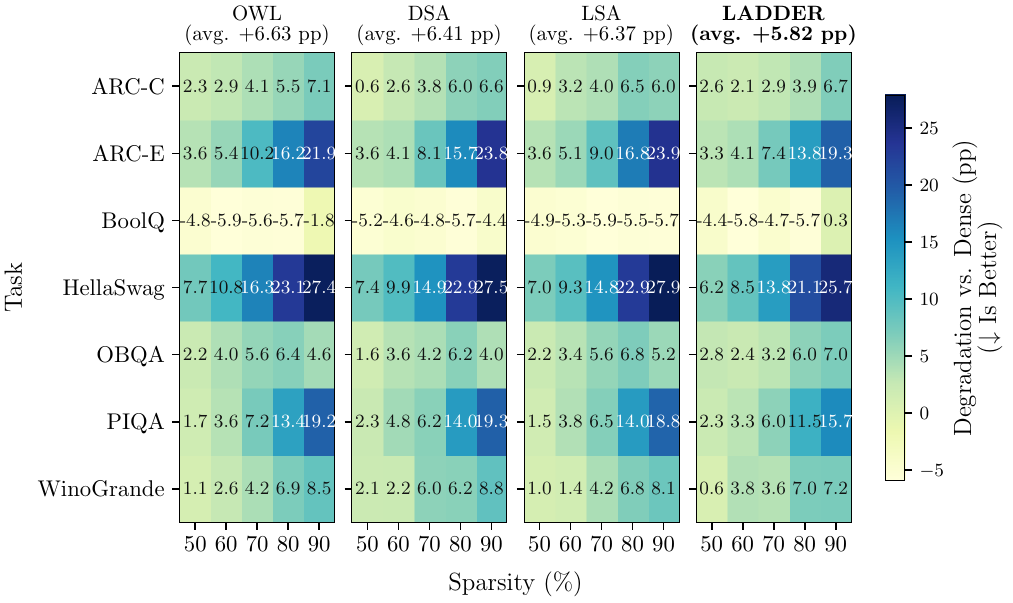}\\[-0.2em]
    {\small (b) After posterior distillation}
    \caption{
        OPT-1.3B zero-shot accuracy degradation before and after posterior distillation.
        Lower is better.
    }
    \label{fig:appendix-opt13-downstream-degradation}
\end{figure*}

\begin{figure*}[t]
    \centering
    \includegraphics[width=0.98\textwidth]{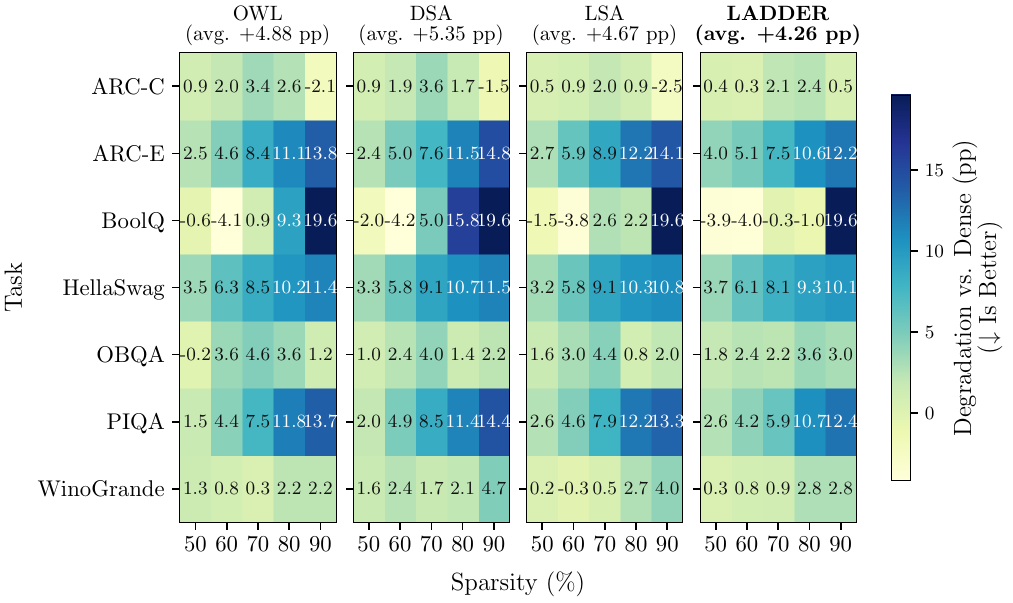}\\[-0.2em]
    {\small (a) Before posterior distillation}\\[0.25em]
    \includegraphics[width=0.98\textwidth]{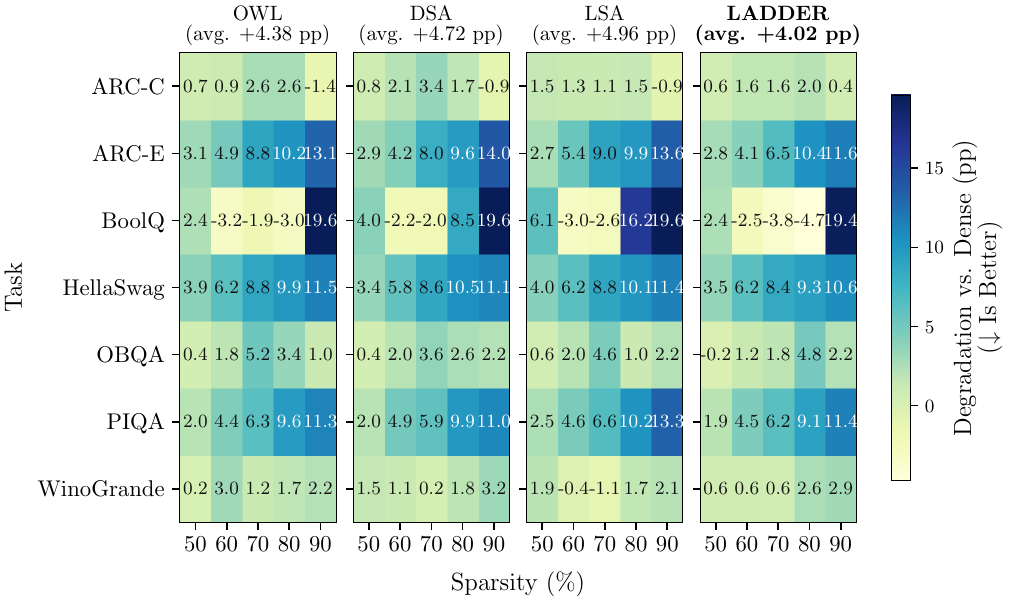}\\[-0.2em]
    {\small (b) After posterior distillation}
    \caption{
        OPT-350M zero-shot accuracy degradation before and after posterior distillation.
        Lower is better.
    }
    \label{fig:appendix-opt350-downstream-degradation}
\end{figure*}

\begin{figure*}[t]
    \centering
    \includegraphics[width=0.98\textwidth]{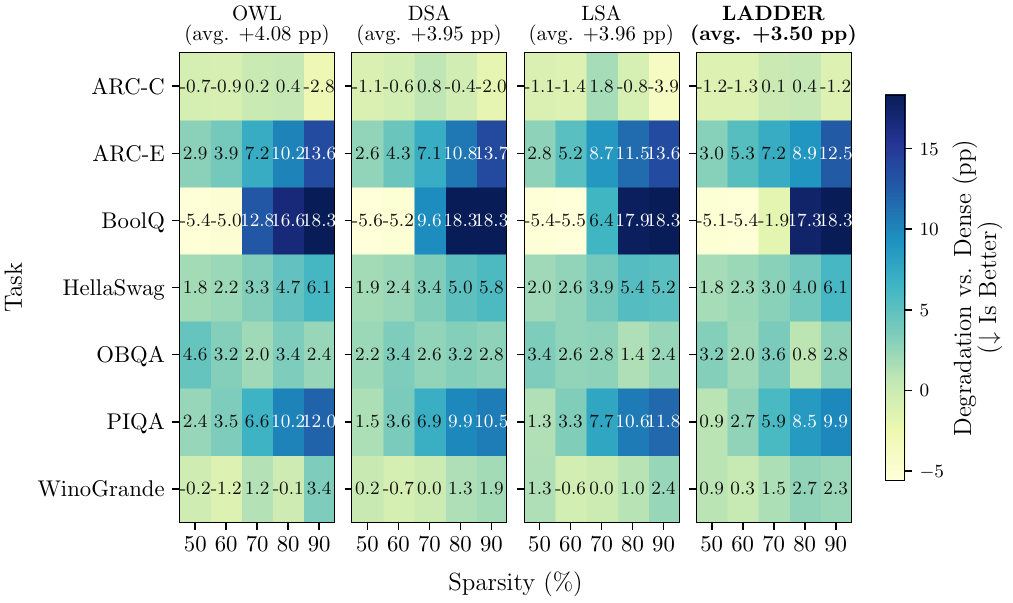}\\[-0.2em]
    {\small (a) Before posterior distillation}\\[0.25em]
    \includegraphics[width=0.98\textwidth]{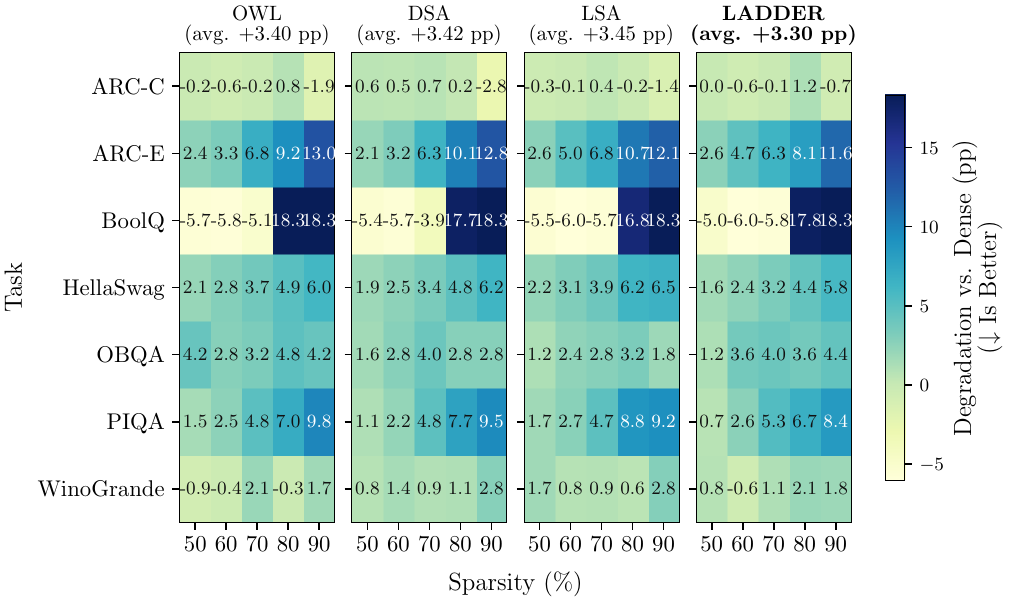}\\[-0.2em]
    {\small (b) After posterior distillation}
    \caption{
        OPT-125M zero-shot accuracy degradation before and after posterior distillation.
        Lower is better.
    }
    \label{fig:appendix-opt125-downstream-degradation}
\end{figure*}

Figures \ref{fig:appendix-qwen-downstream-degradation}--\ref{fig:appendix-opt125-downstream-degradation} show the task-level degradation from each model's dense checkpoint before and after distillation.
Smaller values indicate better accuracy retention.

These paired views preserve variation across tasks that is hidden by the macro average.
The leading method changes across tasks and sparsity levels, with ties and occasional reversals even when the language-modeling losses are clearly separated.
The downstream evidence therefore supports a claim of stronger aggregate accuracy retention at high sparsity, rather than superiority on every task or configuration.

\subsection{Allocation Patterns and Mask Nestedness}
\label{sec:appendix-diagnostics}

\paragraph{Support nestedness.}
Let $M_{\layer}^{(a)}$ and $M_{\layer}^{(b)}$ denote the boolean keep masks for the same tensor at sparsity levels $s_a<s_b$.
The higher-sparsity mask is nested inside the lower-sparsity mask when
$M_{\layer}^{(b)} \subseteq M_{\layer}^{(a)} \Longleftrightarrow M_{\layer}^{(b)} \wedge \neg M_{\layer}^{(a)} = 0$.
The corresponding model-level violation count is $V(a,b) = \sum_{\layer=1}^{\Llayers}\norm{M_{\layer}^{(b)} \wedge \neg M_{\layer}^{(a)}}_0$, with perfect nestedness if and only if $V(a,b)=0$.

\begin{figure*}[t]
    \centering
    \begin{minipage}[t]{0.48\textwidth}\centering
        \includegraphics[width=\linewidth]{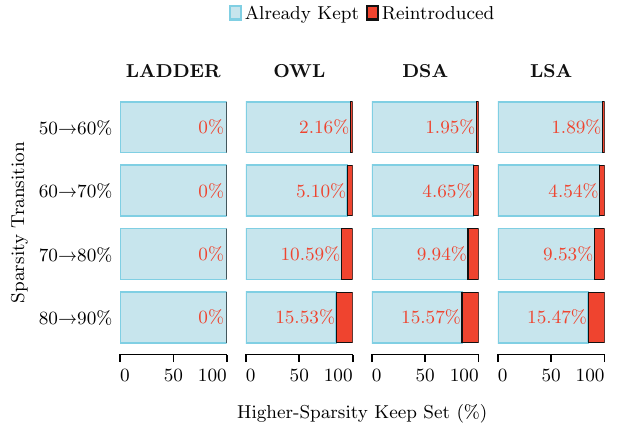}
        \par\smallskip (a) Qwen3-0.6B
    \end{minipage}\hfill
    \begin{minipage}[t]{0.48\textwidth}\centering
        \includegraphics[width=\linewidth]{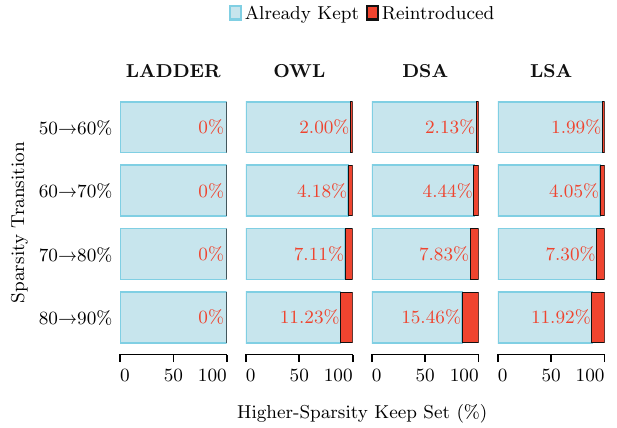}
        \par\smallskip (b) Llama-3.2-1B
    \end{minipage}\\[-0.1em]
    \begin{minipage}[t]{0.48\textwidth}\centering
        \includegraphics[width=\linewidth]{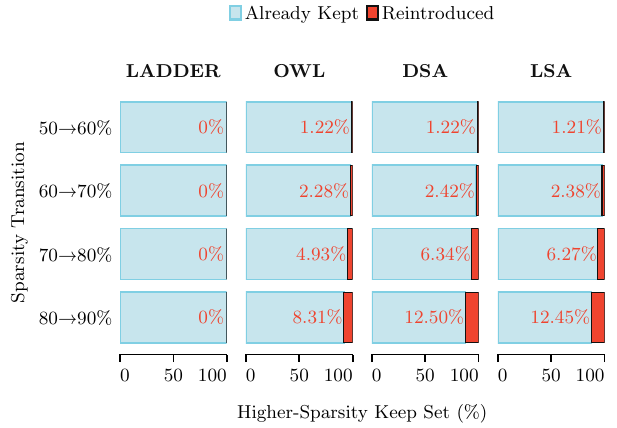}
        \par\smallskip (c) OPT-1.3B
    \end{minipage}\hfill
    \begin{minipage}[t]{0.48\textwidth}\centering
        \includegraphics[width=\linewidth]{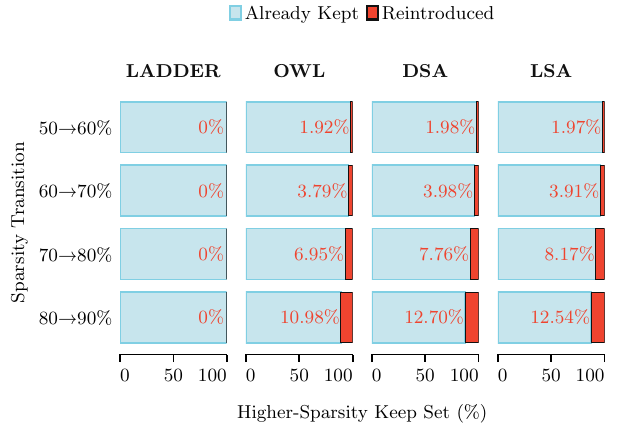}
        \par\smallskip (d) OPT-350M
    \end{minipage}\\[-0.1em]
    \begin{minipage}[t]{0.48\textwidth}\centering
        \includegraphics[width=\linewidth]{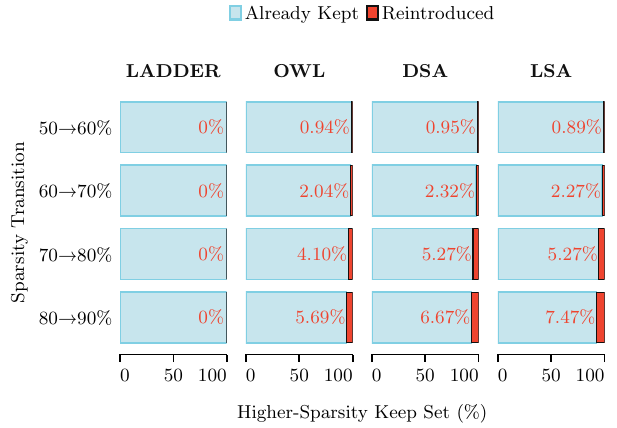}
        \par\smallskip (e) OPT-125M
    \end{minipage}
    \caption{
        Support across adjacent sparsity levels for the five evaluated models. Each bar partitions the weights kept at the higher sparsity into those retained from the next lower sparsity and those reintroduced after having been pruned.
    }
    \label{fig:appendix-model-nestedness}
\end{figure*}

\begin{figure*}[t]
    \centering
    \begin{minipage}[t]{0.48\textwidth}\centering
        \includegraphics[width=\linewidth]{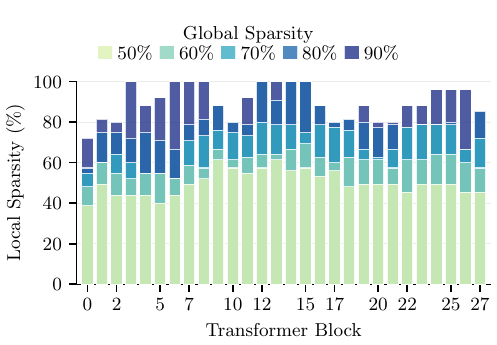}
        \par\smallskip (a) Qwen3-0.6B
    \end{minipage}\hfill
    \begin{minipage}[t]{0.48\textwidth}\centering
        \includegraphics[width=\linewidth]{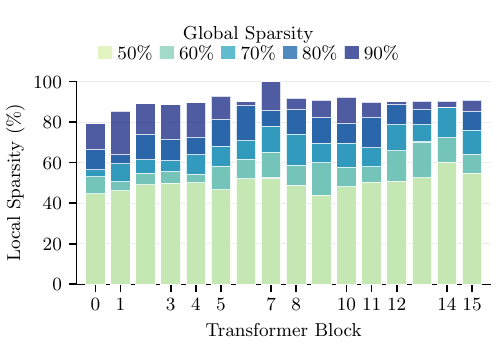}
        \par\smallskip (b) Llama-3.2-1B
    \end{minipage}\\[-0.1em]
    \begin{minipage}[t]{0.48\textwidth}\centering
        \includegraphics[width=\linewidth]{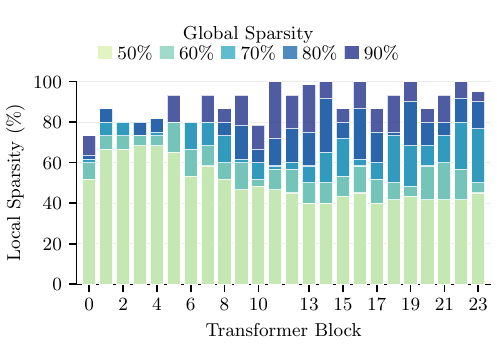}
        \par\smallskip (c) OPT-1.3B
    \end{minipage}\hfill
    \begin{minipage}[t]{0.48\textwidth}\centering
        \includegraphics[width=\linewidth]{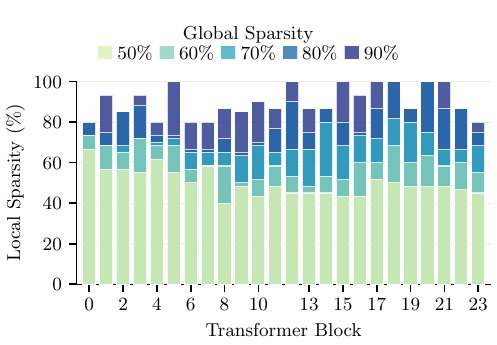}
        \par\smallskip (d) OPT-350M
    \end{minipage}\\[-0.1em]
    \begin{minipage}[t]{0.48\textwidth}\centering
        \includegraphics[width=\linewidth]{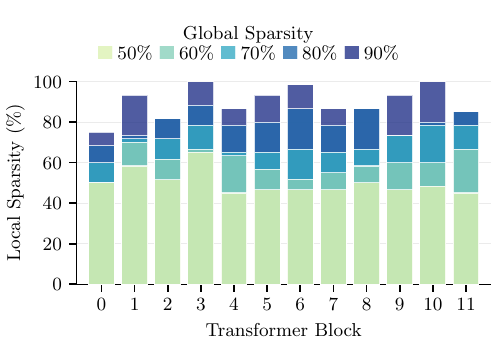}
        \par\smallskip (e) OPT-125M
    \end{minipage}
    \caption{
        Layer allocations selected by \method{} with SparseGPT. Each bar gives the percentage of prunable weights removed from a transformer block; overlaid bars show the allocations from 50\% to 90\% global sparsity.
    }
    \label{fig:appendix-allocation-bars}
\end{figure*}

Across the four measured transitions from 50\% to 90\%, \method{} reintroduces zero weights for every model in Figure \ref{fig:appendix-model-nestedness}. OWL, DSA, and LSA generate each target independently and reintroduce some weights in every measured transition. This comparison verifies mask nestedness; it does not isolate its effect on language-modeling performance. SparseGPT can still update the values of surviving weights after an accepted step, although a pruned weight is never restored.
Figure \ref{fig:appendix-allocation-bars} shows that the selected layer sparsities change with both the model and the global sparsity target. We therefore do not interpret the plots as a fixed ranking of transformer layers.

\end{document}